\pdfoutput=1
\documentclass[11pt]{article}

\usepackage{ACL2023}

\usepackage{times}
\usepackage[ruled]{algorithm2e}
\usepackage{latexsym}
\usepackage[T1]{fontenc}
\usepackage[utf8]{inputenc}
\usepackage{multirow}

\usepackage{microtype}
\usepackage{booktabs}
\usepackage{subcaption}
\usepackage{graphicx}
\usepackage{CJK}
\usepackage{amssymb}
\usepackage{amsmath}
\usepackage{xspace}
\usepackage{enumitem}

\newcommand{\name}{Lego-MT\xspace}

\title{\name: Learning Detachable Models for 
 \\ Massively Multilingual Machine Translation}
\author{
Fei Yuan\textsuperscript{\rm1}, Yinquan Lu\textsuperscript{\rm1}, Wenhao Zhu\textsuperscript{\rm2},  
Lingpeng Kong\textsuperscript{\rm3}, Lei Li\textsuperscript{\rm4}, Yu Qiao\textsuperscript{\rm1}, Jingjing Xu\textsuperscript{\rm1} \\
\textsuperscript{\rm 1} Shanghai Artificial Intelligence Laboratory \\
\textsuperscript{\rm 2} National Key Laboratory for Novel Software Technology, Nanjing University, China \\
\textsuperscript{\rm 3} The University of Hong Kong, \textsuperscript{\rm 4} University of California, Santa Barbara \\
  \texttt{\{yuanfei, luyinquan, qiaoyu\}@pjlab.org.cn},  \texttt{zhuwh@smail.nju.edu.cn} \\ 
   \texttt{lpk@cs.hku.hk},  \texttt{leili@cs.ucsb.edu},  \texttt{jingjingxupku.02@gmail.com}
}

\begin{document}
\maketitle

\begin{abstract}

Multilingual neural machine translation (MNMT) aims to build a unified model for many language directions. 
Existing monolithic models for MNMT encounter two challenges: parameter interference among languages and inefficient inference for large models.
In this paper, we revisit the classic multi-way structures and develop a detachable model by assigning each language (or group of languages) to an individual branch that supports plug-and-play training and inference. To address the needs of learning representations for all languages in a unified space, we propose a novel efficient training recipe, upon which we build an effective detachable model, Lego-MT.
For a fair comparison, we collect data from OPUS and build a translation benchmark covering 433 languages and 1.3B parallel data. 
Experiments show that Lego-MT with 1.2B parameters brings an average gain of 3.2 spBLEU. It even outperforms M2M-100 with 12B parameters. 
The proposed training recipe brings a 28.2$\times$ speedup over the conventional multi-way training method.\footnote{ \url{https://github.com/CONE-MT/Lego-MT}.}

\end{abstract}

\section{Introduction}
% 背景
% multilingual MT的目标
% MMT 的好处
 Multilingual neural machine translation (MNMT) translates languages by mapping a source sentence to a unified representation space and decoding a target sentence from this space~\cite{johnson2017google, gu2018universal, neubig2018rapid, aharoni2019massively, zhang2020improving}. Traditional MNMT models use a shared network to align representations in different languages. Recently, scaling up the size of MNMT models brings significant quantitative improvements and new qualitative capabilities (M2M-100, \citealt{fan2021beyond}; NLLB-200, \citealt{DBLP:journals/corr/abs-2207-04672}; \emph{inter alia}). Beyond MNMT, recent large-scale language models (e.g., ChatGPT) also show promising results on zero-shot (or few-shot) translation, especially for language-to-English translation. Despite great potential, there is still a large gap between LLMs and existing MNMT models on massive translation directions.   %It can maps a source sentence in any language to a common continuous representation space and decodes the representation into any
%of the target languages. Scaling up the size of MNMT models has been a new trend, which brings quantitative improvement and new qualitative capabilities, e.g. M2M-100 (12B) covering 100 languages \cite{fan2021beyond} and NLLB (54.5B) covering 200 languages \cite{costa2022no}
%This manner eases the development of translation systems for many language directions \cite{arivazhagan2019massively} and enables the knowledge transfer among related language pairs \cite{tan2019multilingual}.
% 冲突
% 更多的语言对时，简单scale up不行
%\WH{Despite the great advances of MNMT over the past years 现在的研究遇到什么问题?}, 
%Recently, researchers start to explore the possibility of building a MNMT system for more languages by scaling the model, e.g. M2M (12B) covering 100 languages \cite{fan2021beyond} and NLLB (54.5B) covering 200 languages \cite{costa2022no}.
% But how to effectively and efficiently develop a MNMT system for extreme massive languages (e.g., 400 languages) remains a challenge.
% Previous works  explore this issue mainly by scaling the Transformer model .

\begin{figure}[!t]
    \centering
    \includegraphics[trim={0cm 11.8cm 15cm 1cm},clip,scale=0.50]{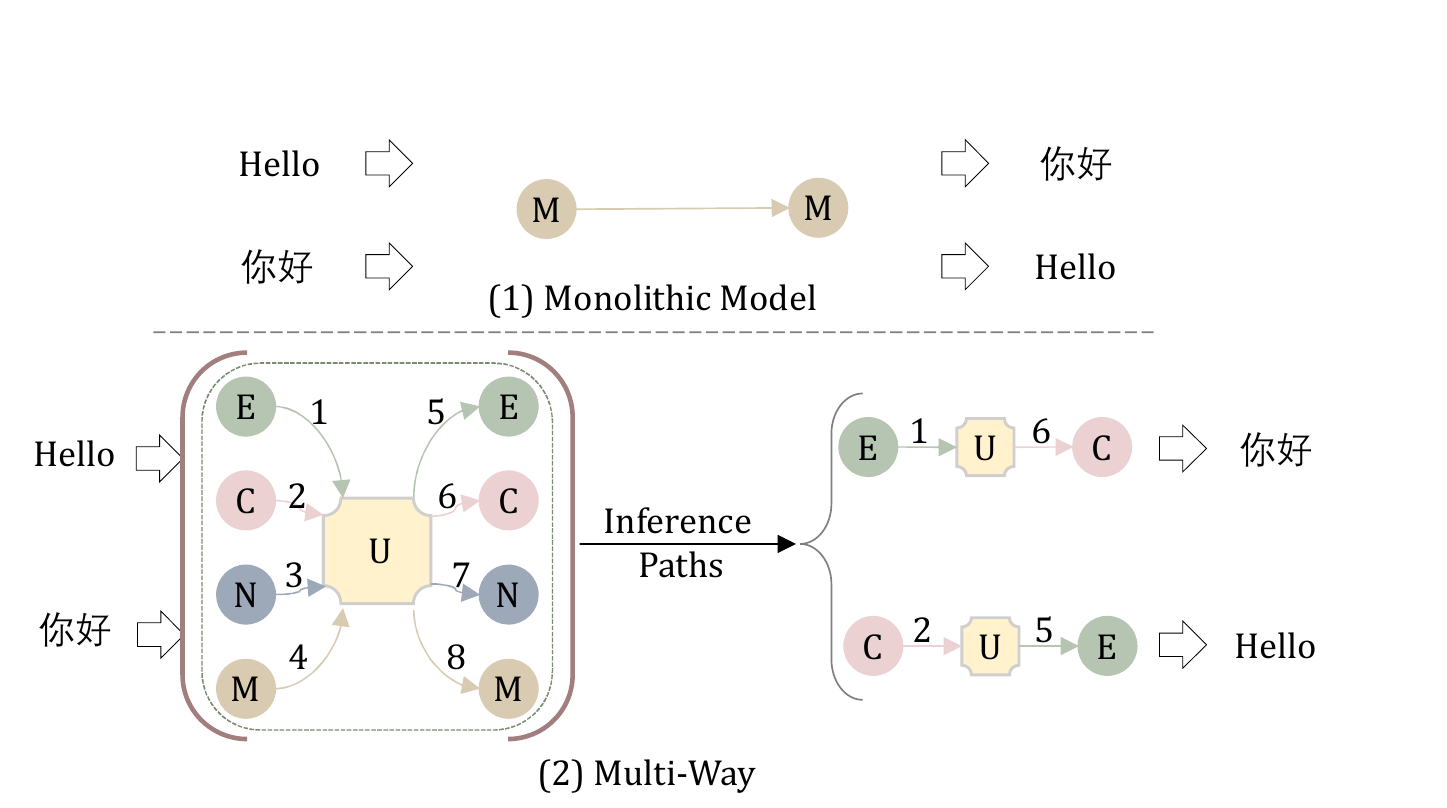}
    \vspace{-8pt}
    \caption{Multi-way architecture.  \emph{(1) Monolithic Model} is the fully-shared model for all translation directions;  \emph{(2) Lego-MT} is a multi-way structure that includes both multilingual (denoted as M) and language-specific encoders and decoders for English (denoted as E), Chinese (denoted as C) and Nepali (denotes as N). The architecture is detachable at inference time where only a specific encoder and decoder are needed. U (Unified space) represents hidden representations generated by encoders.  }
    %, where Language-specific flow refers to a new flow consisting of a language-specific encoder and a language-specific decoder trained by Lego-MT model
    \label{fig:introduce_overview}
\end{figure}

% However, simply using a shared model for massive MNMT brings 
%  new effectiveness and efficiency issues: \textit{1) Parameter Interference}: Memorizing multilingual knowledge within finite parameters cause parameter interference~\cite{ha-etal-2016-toward}, especially between high-resource and low-resource languages \cite{li2021robust}, resulting in significant performance degradation. 
%  \textit{2) High Inference Latency}: The centralization feature requires all parameters to be included in the computation graph during inference, resulting in huge computation costs~\cite{song2021switch}.  Adapter-based approaches~\citep{DBLP:conf/emnlp/ZhuFZWL21} handle parameter interference via fine-tuning new parameters to fit bilingual translation. However, they still fail to adapt massive translation directions. Mixture-of-expert (MoE) support dynamic activation, yet still requires all parameters to be loaded into memory, which does not satisfy our requirements for efficient training and inference.   %Mixture-of-expert (MoE)~\citep{fan2021beyond,DBLP:journals/corr/abs-2207-04672} also provides a potential solution by dynamically activating model parameters for any input. %However, MoE still requires all parameters to be loaded in GPU memory, which does not fit the requirements for efficient training and inference. 
 
%\end{enumerate}
%Due to these issues, how to effectively and efficiently develop a massive MNMT system remains a challenge.
% where the catastrophic forgetting \cite{mccloskey1989catastrophic} problem usually occurs and .
% our method

Simply using a shared model for massive MNMT brings 
 new effectiveness and efficiency issues. First, memorizing multilingual knowledge within finite parameters causes parameter interference~\cite{ha-etal-2016-toward}, especially between high-resource and low-resource languages \cite{li2021robust}, which leads to significant performance degradation.
 Second, the centralization feature requires all parameters to be included in the computation graph during the inference stage, resulting in heavy computational overhead ~\cite{song2021switch}.
 % Adapter-based approaches~\citep{DBLP:conf/emnlp/ZhuFZWL21} handle parameter interference via fine-tuning new parameters to fit bilingual translation. However, they still fail to adapt massive translation directions. Mixture-of-expert (MoE) support dynamic activation, yet requires all parameters to be loaded into memory, which does not satisfy our requirements for efficient training and inference.
 Common fixes of these issues include adapter-based approaches~\citep{DBLP:conf/emnlp/ZhuFZWL21}, which handle parameter interference via fine-tuning new parameters to fit bilingual translation, and mixture-of-expert (MoE), which supports dynamic activation. These methods either fail to adapt to massive translation directions or require all parameters to be loaded into memory, thus remaining unsatisfactory considering the efficiency of training and inference.

To find out the best recipe for massive multilingual translation, we revisit the classic multi-way (or multi-branch) architecture~\cite{dong2015multi,firat2016multi}, whose philosophy is to allocate an individual encoder and decoder for each language (or group of languages), as shown in Figure~\ref{fig:introduce_overview}. The immediate benefit of this structure is: 1) The utilization of individual modules for specific languages mitigates parameter interference; 2) Each branch can be independently loaded during inference, significantly reducing computational costs and decreasing inference latency.
% Considering these strengths, we aim to build an effective detachable multi-way model. 

Despite appealing, there remain two big challenges when training multi-way structures:  \textit{representation alignment} between different languages due to the lack of shared parameters;  and  \textit{low GPU efficiency} during training because unused parameters occupy GPU memory but do not have any computations. Furthermore, the feature of random language mixture in a batch makes it infeasible to use an online-loading method (i.e., loading during usage) to accelerate training since it will cause impractical IO communication costs during batch switching (between CPU and GPU). 

To address these challenges, we propose a novel training recipe, which results in our new detachable model, Lego-MT. 
%To make multi-branch training more scalable and efficient, we propose two techniques and present a novel architecture and a new training framework, Lego-MT.
% we propose a novel training framework, Lego-MT, where  is introduced to fix the flaw.
%First, we design a localized training strategy. 
We classify the training data into different language-centric groups such that we only need to load specific branches into GPU memory, eliminating the need to load different modules constantly. The language-centric group is trained in sequential order. 
Second, during each language-centric training, we introduce a multilingual branch and propose a new triple-flow method to help a model learn to map to and translate from a unified space. Specifically, a unified space is a type of representation space rather than a module. It creates a common representation of language that can be used across multiple language tasks.

%Specifically, we first classify multilingual data into language-specific groups and cut the whole de-centralized architecture into several branch pieces. The language-specific group is trained in sequential order where only related branch pieces are loaded when training a language-specific group. %This loop keeps running until we reach the maximum number of loops.
%map the representation of different languages into a unified semantic space to encourage knowledge transferring and lessen catastrophic forgetting \cite{mccloskey1989catastrophic}.
%Intuitively, the alignment enables the encoder and decoder belongs to different languages to better cooperate with each other, even if the given translation direction is unseen during training.

% experiments

To evaluate our training recipe for massive MNMT, we construct a many-to-many translation dataset\footnote{The dataset is released on \url{https://github.com/CONE-MT/Lego-MT.git}.} covering 7 language-centric groups, 433 languages, based on the open-source website OPUS\footnote{\url{https://opus.nlpl.eu}.} \cite{tiedemann2012parallel}. 

Lego-MT-1.2B yields average gains of 3.2 spBLEU, and even outperforms M2M-100-12B which has 10$\times$ inference parameters. %Fine-tuning M2M-100-1.2B on the same training data only brings slight improvements, 1.8 spBLEU on many-to-one translation and -0.1 spBLEU on one-to-many translation. 
Furthermore, the proposed training recipe brings a 28.2$\times$ speedup compared with the conventional multi-way training method.
% Due to the detachable features, 
We also conduct comprehensive experiments on branch combinations, thanks to the detachable nature of the model. We find that low-resource languages prefer multilingual branches and high-resource languages prefer language-specific branches. 
%With specific branch combinations, we can get higher improvements.
In addition, we also observe that the unseen combination of a high-resource language encoder and a high-resource language decoder can achieve better performance, showing that Lego-MT can align different branches into a unified space effectively. 
The main contributions can be summarized as follows:

\begin{itemize}[nosep,leftmargin=1em]
    \item We build an effective detachable model Lego-MT for multilingual machine translation.
    \item Experiments demonstrate that Lego-MT brings an average gain of 3.2 spBLEU. This training recipe results in a 28.2$\times$ training speedup compared with the naive multi-branch architecture.
    \item We construct a massive multilingual translation dataset covering 433 languages, which greatly extends the scale of languages.
\end{itemize}

\begin{figure*}[h]
    \centering
    \includegraphics[trim={0cm 5.8cm 1cm 2.7cm},clip,scale=0.48]{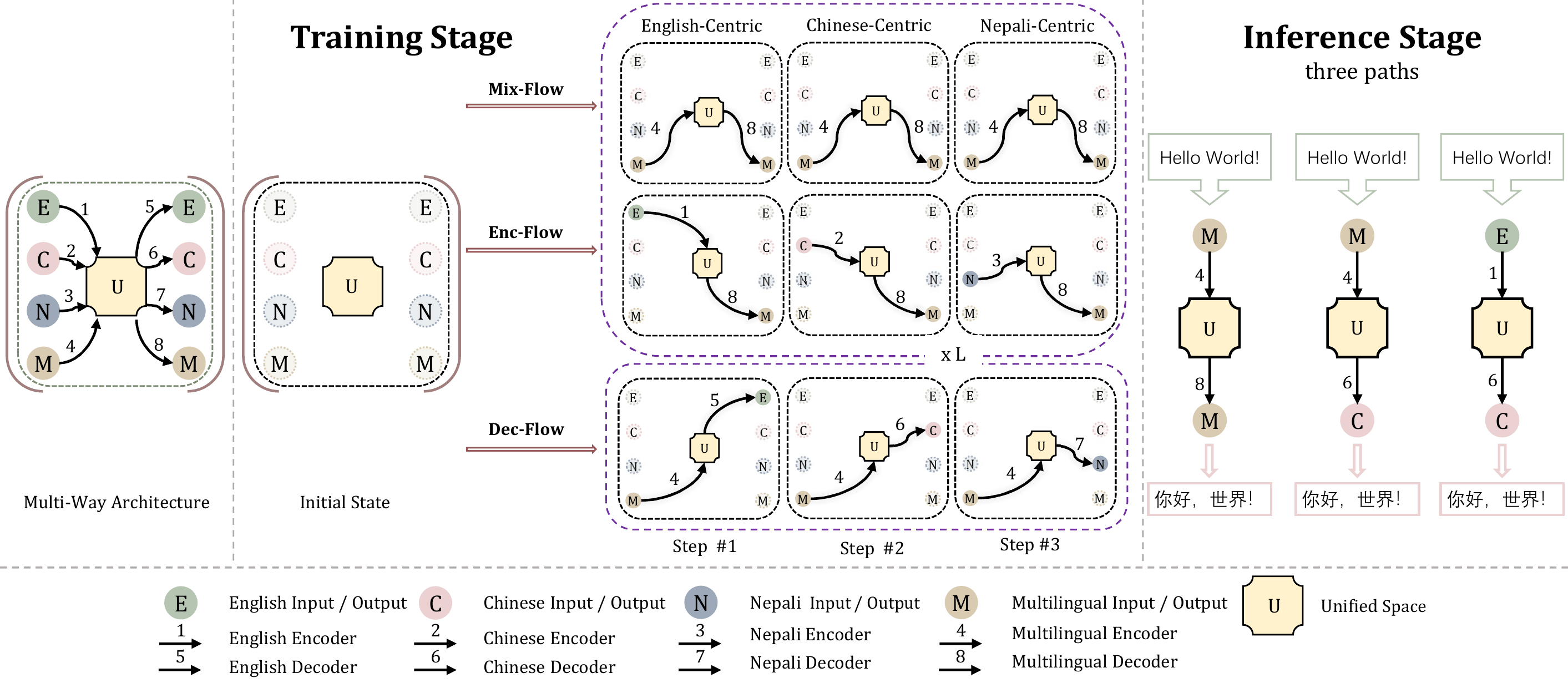}
    \caption{The overview of Lego-MT and training recipe. During training, we introduce an efficient training method by classifying multilingual data into language-centric groups.  The language-centric groups are trained in a sequential way. During each training phase, only  language-specific parameters are loaded into GPU memory. The training maintains three flows: \textit{Enc-Flow} (language-specific encoder + multilingual decoder) for training specific encoder, \textit{Dec-Flow} (Multilingual encoder + language-specific decoder) to train language-specific decoder, and \textit{Mix-Flow} (multilingual encoder + multilingual decoder) to avoid the overfitting of multilingual encoder and decoder to each language-centric training data. U means the unified space, hidden representations generated by encoders.  }
    \label{fig:overivew}
\end{figure*}

\section{Related Work}
In this part, we review recent related multilingual machine translation models. We classify them into three categories: fully / group-shared~\cite{survey-mnmt}, and Mixture-of-expert (MoE). 

The fully-shared model is the most prevalent model in Multilingual Neural Machine Translation (MNMT). This model employs a single architecture to translate in all directions~\cite{2017-1-1-model, johnson2017google, bapna2019simple, lin2020pre, liu2020multilingual, pan2021contrastive, sun2021multilingual} and has demonstrated efficacy in aiding low-resource directions. However, fully-shared models are often subject to capacity bottlenecks and trade-offs between translation quality and the number of languages~\cite{aharoni2019massively,zhang2020improving,ha-etal-2016-toward}. Group-shared models incorporate individual parameters for each group and represent a popular solution for sharing language-specific encoders or decoders~\cite{lee2017fully,zoph2016multi}. \citet{lee2017fully,sachan-neubig-2018-parameter,ji2020cross,lyu-etal-2020-revisiting, sachan-neubig-2018-parameter} proposed an MNMT model only for shared language-specific modules. 
LaSS~\cite{lin2021learning} learns language-specific sub-networks for each language direction for multilingual translation. 
Adpater methods~\cite{bapna-firat-2019-simple,DBLP:conf/emnlp/ZhuFZWL21} add additional side networks to each language direction in addition to the main multilingual Transformer encoder-decoder. 
While these studies can alleviate the capacity bottleneck to some extent, challenges remain when handling larger-scale languages.

Mixture-of-Expert (MoE) has recently emerged as a prominent research direction~\cite{jacobs1991adaptive,shazeer2017outrageously,lepikhin2020gshard,fedus2021switch,du2022glam,fan2021beyond,DBLP:journals/corr/abs-2207-04672}, which are sparsely activated, with each inference only activating a subset of parameters. Researchers have applied MoE to massively multilingual translation and introduced various regularization strategies to enhance performance~\cite{DBLP:conf/acl/Dai0MZSCW22,DBLP:journals/corr/abs-2207-04672}. Despite promising results, MoE’s objective differs from ours, as it still requires the entire structure to be stored in GPU memory during inference.

The encoder-decoder structure has demonstrated considerable flexibility through the utilization of the Lego-NN~\cite{dalmia2022legonn}. The Lego-NN can be applied to various tasks with decoder modules being detachable, in contrast, the Lego-MT model design allows for the performance of massively MNMT with \textbf{all modules} being detachable.

\section{Lego-MT}
\label{sec:approach}

%%%%%%%%%%%%%%%%%%%%%%%%%%%%%%%%%%%%%%%%%%%%%%%%%%
\begin{algorithm*}[t]
\footnotesize
\KwIn{Epoch number $L$. Training data for Mix-Flow, Enc-Flow and Dec-Flow: $\mathcal{D}_{\mathrm{multi}} = \{\mathcal{D}_{s_1 \rightarrow t_1}, \mathcal{D}_{s_i \rightarrow t_j},..., \mathcal{D}_{s_N \rightarrow t_N}\}$ and  $\mathcal{D}_{\mathrm{lg}\rightarrow \cdot} = \{\mathcal{D}_{\mathrm{lg} \rightarrow t_1}, \mathcal{D}_{\mathrm{lg} \rightarrow t_j},..., \mathcal{D}_{\mathrm{lg} \rightarrow t_N}\}$ and $\mathcal{D}_{\cdot \rightarrow \mathrm{lg}}=\{\mathcal{D}_{s_1 \rightarrow \mathrm{lg}}, \mathcal{D}_{s_i \rightarrow \mathrm{lg}},..., \mathcal{D}_{s_N  \rightarrow \mathrm{lg}}\}$, respectively. The parameters used for Mix-Flow and Enc-Flow are initialized as $\theta_m=\theta_0$ and $\theta_e=\theta_0$. Note, the parameters used for Dec-Flow are initialized as $\theta_d=\theta_m$ after training of Mix-Flow and Enc-Flow. } 

\For {epoch $l = 1$ to $L$}   
{
	Shuffle $\mathcal{D}_{\mathrm{lg}\rightarrow \cdot}$ to obtain a new training sequence. \\
	\For {each batch $\mathcal{D}_e \in \mathcal{D}_{\mathrm{lg}\rightarrow \cdot}$} 
	{
        Evaluate the objective by Equation~\ref{eq:l_e} on $\mathcal{D}_e$: 
        $l_e = \sum_{\mathbf{x,y}\sim\mathcal{D}_b} - \mathrm{log}P_{\theta_e}(\mathbf{y}|\mathbf{x})$  \\
	    Get a minibatch of multilingual data $\mathcal{D}_m \in \mathcal{D}_{\mathrm{multi}}$ \\
        Evaluate the objective by  Equation~\ref{eq:l_m}  on $\mathcal{D}_m$: 
        $l_m = \sum_{\mathbf{x,y}\sim\mathcal{D}_m} - \mathrm{log}P_{\theta_m}(\mathbf{y}|\mathbf{x}) $\\
        Update $\theta_m$ and $\theta_e$ by: 
        $\theta_m \leftarrow \theta_m - \eta \bigtriangledown_{\theta_m}{(l_m+l_e)} $ and $\theta_e \leftarrow \theta_e - \eta \bigtriangledown_{\theta_e}{l_e} $
        
	}
}
 \For {epoch $l = 1$ to $L$}   
{
	Shuffle $\mathcal{D}_{\cdot \rightarrow \mathrm{lg}}$ to obtain a new training sequence. \\
	\For {each batch $\mathcal{D}_d \in \mathcal{D}_{\cdot \rightarrow \mathrm{lg}}$} 
	{
        Calculate $\mathcal{D}_d$ by  Equation~\ref{eq:l_d} : 
        $l_d = \sum_{\mathbf{x,y}\sim\mathcal{D}_d} - \mathrm{log}P_{\theta_d}(\mathbf{y}|\mathbf{x})$  \\
        Update $\theta_d$: 
        $\theta_d \leftarrow \theta_d - \eta \bigtriangledown_{\theta_d}{l_d} $
        
	}
}
\caption{Triple-flow training.}
\label{alg: Lego-MT-training}
\end{algorithm*}
%%%%%%%%%%%%%%%%%%%%%%%%%%%%%%%%%%%%%%%%%%%%%%%%%%

% The overview of our model structure, training recipe, and inference procedure is shown in Figure~\ref{fig:overivew}.  First, we describe the whole structure of the Lego-MT and the overview of the training recipe in Section~\ref{sec:structure},  then introduce the details of triple-flow training in Section~\ref{sec:data_flow};  and present the full training and inference details in Section~\ref{sec:procedure}.  

\subsection{Overview}
\label{sec:structure}

This paper aims to build a detachable multi-branch model with a language (or group)-specific encoder and a language (or group)-specific decoder. As shown in Figure~\ref{fig:overivew}, the detachable structure provides an effective mechanism to only load a part of modules during training and inference. %Unlike the classical multi-way design approach, the Lego-MT model concentrates on efficiency and effectiveness, which focuses on mapping the different languages representation into a unified space and can be disassembled and assembled optionally.

During training, we introduce a new training method by classifying multilingual data into language-centric groups.  During each training phase, only language-centric data and related branches are loaded. All language-centric groups are trained in a sequential way.  We empirically found that the orders contribute little to the final performance and we fix the training order for simplification in the next parts. 

During each language-centric training phase, we introduce a multi-lingual branch to help language-specific branches learn to encode to a unified space and decode from a unified space. \textbf{Unified Space} is a concept that aims to map all languages into a unified representation space without any parameters. This concept is used in natural language processing and machine learning to create a common representation of language~\cite{lyu-etal-2020-revisiting, fan2021beyond} that can be used across different languages.

The training maintains triple-flow: \textit{Enc-Flow} (language-specific encoder + multilingual decoder) for training specific encoder, \textit{Dec-Flow} (multilingual encoder + language-specific decoder) to train language-specific decoder, and \textit{Mix-Flow} (multilingual encoder + multilingual decoder) to avoid the overfitting of multilingual encoder and decoder to each language-centric training data. 
Surprisingly, we find that Dec-flow cannot be trained together with Mix/Enc-flow, resulting in catastrophic forgetting in the multilingual encoder (detailed discussion in Section~\ref{sec:branch_ablation}). Therefore, the basic training processes can be briefly divided into two stages: the Mix/Enc-Flow phase and the Dec-Flow phase.  %This is principally due to the negative effects between M-Flow and D-Flow

During inference, there are three alternative flows in Lego-MT for  language-centric translation to be translated (``Inference Stage'' in Figure~\ref{fig:overivew}). %This is mainly because each component of the Lego-MT model is aligned on the unified space during the training. Therefore, a certain translation pair has multiple paths. 
As shown in Figure~\ref{fig:overivew}, users can decide to choose which path for inference.  %English sentences are not only the input of the English encoder but also the input of the multilingual encoder. The same is true for Chinese sentences in Figure~\ref{fig:overivew}. The best combination of Lego-MT modules will be discussed in more detail in a later section~\ref{sec:exp_res}.

\subsection{Triple-Flow Training}
\label{sec:data_flow}
%Three challenges in training that the Lego-MT model is facing are 1)  balancing the cross-lingual information sharing with the parameter competition and 2) catastrophic forgetting caused by sequential training. To this end, we introduce triple data flow in training (as "Training Stage" shown in Figure~\ref{fig:overivew}). 

Given a multilingual dataset with ${N}$ languages, $\mathcal{D}_{\mathrm{multi}} = \{\mathcal{D}_{s_1 \rightarrow t_1}, \mathcal{D}_{s_i \rightarrow t_j},..., \mathcal{D}_{s_N \rightarrow t_N}\}$, where each $\mathcal{D}_{s_i \rightarrow t_j}$ contains a parallel data from the source language $S_i$ to the target language $T_j$, $s_i$ refers to the $i$-th ($i\in N$) language  being translated from, $t_j$ represents the  $j$-th ($j\in N$) language being translated into, respectively. %$0$ is allowed for dataset size of $\mathcal{D}_{s_i \rightarrow t_j}$. 
Specifically, one-to-many multilingual data for a specific language ($\mathrm{lg}$) can be expressed as $\mathcal{D}_{\mathrm{lg}\rightarrow \cdot} = \{\mathcal{D}_{\mathrm{lg} \rightarrow t_1}, \mathcal{D}_{\mathrm{lg} \rightarrow t_j},..., \mathcal{D}_{\mathrm{lg} \rightarrow t_N}\}$. Similarly, the many-to-one multilingual data for a specific language ($\mathrm{lg}$) can be denoted as $\mathcal{D}_{\cdot \rightarrow \mathrm{lg}}=\{\mathcal{D}_{s_1 \rightarrow \mathrm{lg}}, \mathcal{D}_{s_i \rightarrow \mathrm{lg}},..., \mathcal{D}_{s_N  \rightarrow \mathrm{lg}}\}$. All input sequence is preceded by a special tag (called the language tag) to indicate the source language and target languages. During each training phase, we have triple-flows playing for different rules, Mix-Flow, Dec-Flow, and Enc-Flow.

\subsubsection{Mix-Flow}

Mix-Flow is built upon a multilingual encoder branch and a multilingual decoder branch. It is trained on multilingual to multilingual data. This flow learns a mapping function $f$ from a sentence in any language to another language.  All language  data is mixed together. The input source sequence is preceded by a special tag (called the language tag) to indicate the source languages. Following traditional methods, we also add a target language tag in the decoder part.  The training loss for a Mix-Flow is:
\begin{equation}
    \label{eq:l_m}
    \mathcal{L}_m = -\sum_{\mathbf{x,y}\sim\mathcal{D}_{\mathrm{multi}}} \mathrm{log}P_{\theta_m}(\mathbf{y}|\mathbf{x})
\end{equation}
where $\mathbf{x},\mathbf{y}$ is a pair sampled from multilingual training data. It is used to avoid over-fitting language-specific data in Enc-Flow and Dec-Flow. 

\subsubsection{Enc-Flow}
%Note, the parameters of the multilingual decoder can be  affected by the language-specific encoder.
Enc-Flow includes a language-specific encoder and a multilingual decoder. It is trained with one-to-many multilingual data.  The structure of such a design is natural for language-specific encoder training: the encoder input data comes from the same source language $\mathrm{lg}$, and the decoder is multi-lingual data. The language tag is also added to the encoder and decoder parts. The training loss for language-specific Enc-Flow is:
\begin{equation}
    \label{eq:l_e}
    \mathcal{L}_e = -\sum_{\mathbf{x,y}\sim\mathcal{D}_{\mathrm{lg}\rightarrow \cdot}} \mathrm{log}P_{\theta_e}(\mathbf{y}|\mathbf{x})
\end{equation}
where $\mathbf{x}, \mathbf{y}$ is a pair sampled from one-to-many training data.

\subsubsection{Dec-Flow}
Dec-Flow includes a multilingual encoder and a language-specific decoder. It is trained with many-to-one translation. We separate the training of Dec-Flow from the training of Enc-Flow and Mix-Flow. The parameters used for training Dec-Flow are initialized with the latest model trained by Mix-Flow and Enc-Flow. The language tag is also added to the encoder and decoder parts.  Given a many-to-one dataset $\mathcal{D}_{\cdot \rightarrow \mathrm{lg}}$, the training loss is:
\begin{equation}
    \label{eq:l_d}
    \mathcal{L}_d = -\sum_{\mathbf{x,y}\sim\mathcal{D}_{\cdot \rightarrow \mathrm{lg}}} \mathrm{log}P_{\theta_d}(\mathbf{y}|\mathbf{x})
\end{equation}
where $\mathbf{x}, \mathbf{y}$ is a pair sampled from many-to-one training data. 

% \paragraph{Discussion with Adapter}

\subsection{Training Algorithm}
\label{sec:procedure}

Algorithm~\ref{alg: Lego-MT-training} shows the whole training procedure. We will go into the effects of the two-stage design in Section~\ref{sec:branch_ablation}. In the first stage, we initialize each module of  the Lego-MT model with a pre-trained MT model $\theta_0$. After initialization,  we shuffle a one-to-many dataset to obtain a new training sequence for Enc-Flow training. %Significantly, the number of batches in the M-Flow is consistent with that in the E-Flow, which is far less than the batch number in standard multilingual training.  
In the second stage, we fix the encoder parameter of M-Flow $\theta_m$ and learn the D-Flow decoder $\theta_d$. The iteration keeps running for $L$ epochs. During inference, users can decide to load which flow for inference. We also evaluate the gap between these inference flows in experiments. 

\section{Experiments}

While Lego-MT is generic, we focus the experiments on M2M-100-1.2B as backbone models since M2M-100 is a leading MT model.

% \begin{figure}[t]
%     \centering
%     \includegraphics[scale=0.3]{image/top30_langs4.png}
%     % \includegraphics[height=3cm, width=8cm]{image/top30_langs4.png}
%     \caption{The Size of The Top-20 Languages.}
%     \label{fig:Top-20}
% \end{figure}

\subsection{Dataset}
%OPUS is an open corpus that collects numerous parallel sentences from the web. The OPUS corpus covers a large amount of domains from legislative texts to religious texts. OPUS uses several pre-processing and alignment tools to mine all collected texts and has no manual corrections.

\noindent\textbf{Training Data} We create a Many-to-Many dataset from OPUS$\footnote{\url{https://opus.nlpl.eu/}}$. We build a dataset covering 7 language-specific data and 433 languages. The 7 core languages are En, Zh, De, Ar, Ne, Az, Ceb. The specifics of the construction process are delineated in Appendix~\ref{sec:construction}. All training pairs have been deduplicated with \textit{Flores-101}. %The OPUS corpus covers many domains from legislative to religious texts. We give more details about how to establish our dataset in the following paragraphs.

\noindent\textbf{Evaluation Data} We use \textit{Flores-101}~\cite{fan2021beyond} as the evaluation set, which provides human-written translation pairs covering 101 languages. Since M2M-100 baselines only cover 86 languages, we only compare Lego-MT with baselines on 86 languages\footnote{These 86 languages are: af, am, ar, ast, be, bg, bn, bs, ca, ceb, cs, cy, da, de, el, en, es, et, fa, ff, fi, fr, ga, gl, gu, ha, he, hi, hr, hu, hy, id, ig, is, it, ja, jv, ka, kk, km, kn, ko, lb, lg, ln, lo, lt, lv, mk, ml, mn, mr, ms, my, ne, nl, no, ns, oc, or, pa, pl, ps, pt, ro, ru, sd, sk, sl, so, sr, sv, sw, ta, th, tl, tr, uk, ur, uz, vi, wo, xh, yo, zh, zu.}. We evaluate 7$\times$85 translation directions in total.

%%%%%%%%%%%%%%%%%%%%%%%%%%%%%%%%%%%%%%%%
\begin{table*}[t]
\centering
\small
\resizebox{\textwidth}{!}{
\begin{tabular}{l|c|ccccccc|c}
\toprule
\textbf{Model} & \textbf{Param.} & \textbf{X $\rightarrow$ En}  & \textbf{X $\rightarrow$ Zh} & \textbf{X $\rightarrow$ De} & \textbf{X $\rightarrow$ Ar} & \textbf{X $\rightarrow$ Ne} & \textbf{X $\rightarrow$ Az} & \textbf{X $\rightarrow$ Ceb} & \textbf{AVG.}\\
\midrule
1: Flores-175M \citep{goyal2022flores} & $\times$ 0.1 & 15.7 & 7.2 & 11.2 & 4.6 & 0.6 & 3.0 & 3.1 & 6.5  \\
2: M2M-100-418M \citep{fan2021beyond} & $\times$ 0.3 & 21.2 & 10.3 & 14.2 & 11.5 & 1.3 & 2.4 & 4.9 & 9.4 \\
3: Flores-615M \citep{goyal2022flores} & $\times$ 0.5 & 21.6 & 11.0 & 16.1 & 8.8 & 1.0 & 4.7 & 5.3 & 9.8 \\
4: M2M-100-1.2B \citep{fan2021beyond} & $\times$ 1.0 & 26.3 & 12.9 & 19.3 & 8.1 & 1.4 & 4.6 & 6.8 & 11.3 \\
5: M2M-100-12B \citep{fan2021beyond} & $\times$ 10.0  & 28.0 & 13.3 & 21.3 & 15.1 & 2.9 & 6.4 & 8.8 & 13.7 \\
\midrule
6: (4) + LG-Centric Fine-Tuning  & $\times$ 1.0 & 27.9 & 13.0 & 19.5 & 17.2 & 5.5 & 4.2 & 0.5 & 12.5 \\
7: (4) + Multilingual Fine-Tuning & $\times$ 1.0 & 27.4 & 13.9 & 20.9 & 15.2 & 12.1 & 9.4 & 10.3 & 15.6 \\
\midrule
8: Lego-MT & $\times$ 1.0 & \textbf{30.7} & \textbf{16.4} & \textbf{23.8} & \textbf{18.2} & \textbf{15.0} & \textbf{11.9} & \textbf{15.1} & \textbf{18.7} \\

\midrule
\midrule

\textbf{Model} & \textbf{Param.} & \textbf{En $\rightarrow$ X}  & \textbf{Zh $\rightarrow$ X} & \textbf{De $\rightarrow$  X} & \textbf{Ar $\rightarrow$ X} & \textbf{Ne $\rightarrow$ X} & \textbf{Az $\rightarrow$ X} & \textbf{Ceb $\rightarrow$  X} & \textbf{AVG.}\\
\midrule
1: Flores-175M \citep{goyal2022flores} & $\times$ 0.1 & 12.7 & 7.8 & 11.6 & 6.9 & 2.2 & 2.8 & 5.4 & 7.1 \\
2: M2M-100-418M \citep{fan2021beyond} & $\times$ 0.3 & 17.3 & 10.1 & 14.1 & 11.5 & 4.0 & 4.2 & 6.1 & 9.6 \\
3: Flores-615M \citep{goyal2022flores} & $\times$ 0.5 & 18.0 & 11.1 & 15.6 & 11.2 & 5.2 & 4.3 & 7.9 & 10.5 \\
4: M2M-100-1.2B \citep{fan2021beyond} & $\times$ 1.0 &  21.5 & 13.1 & 17.7 & 12.6 & 7.1 & 6.1 & 9.5 & 12.5 \\
5: M2M-100-12B \citep{fan2021beyond} & $\times$ 10.0  & 24.7 & 14.9 & 20.3 & 16.4 & 9.7 & 6.2 & 12.5 & 15.0\\
\midrule

6: (4) + LG-Centric Fine-Tuning  & $\times$ 1.0 &  21.3 & 10.9 & 15.8 & 14.9 & 3.9 & 3.0 & 1.5 & 10.2 \\
7: (4) + Multilingual Fine-Tuning & $\times$ 1.0 & 21.8 & 13.5 & 18.4 & 14.7 & 13.4 & 11.1 & 12.4 & 15.0\\

\midrule
8: Lego-MT & $\times$ 1.0 & \textbf{25.0} & \textbf{16.3} & \textbf{21.4} & \textbf{18.4} & \textbf{17.0} & \textbf{13.5} & \textbf{16.8} & \textbf{18.3} \\
\bottomrule

\end{tabular}}
\caption{Translation results on \textit{Flores-101}. The top group shows the results of many-to-one translation and the bottom part shows the results of one-to-many settings. We display the spBLEU on the devtest of \textit{Flores-101}. Each cell represents the average performance of translating from the rest languages. ``Param.'' represents the number of required parameters during inference. Baseline 7 has the exact same training data with Lego-MT. For a fair comparison, we use Mix-Flow in Lego-MT for all translation pairs. Lego-MT outperforms M2M-100-1.2B w. multilingual fine-tuning by a large margin, with an average gain of 3.2 spBLEU (3.1 on many-to-one translation and 3.3 on one-to-many translation). }
\label{tab:many-to-one}
\end{table*}

%%%%%%%%%%%%%%%%%%%%%%%%%%%%%%%%%%%%%%%%%

\subsection{Baselines}
\label{sec:model}
We conduct experiments by using a pre-trained multilingual machine translation model: M2M-100-1.2B~\cite{fan2021beyond} as initialization. We build 7 language-specific encoders and 7 language-specific decoders to model 7 core languages. We compare Lego-MT with the following baselines.

\noindent\textbf{Flores-175MB / 615MB} \textit{Flores-101}~\cite{goyal2022flores} furnishes two baseline models, with parameter sizes of 175MB and 615MB respectively, constructed on M2M-100.

\noindent\textbf{M2M-100-418M} It is the smallest model released by \citet{fan2021beyond}, which is a base-version Transformer model with 12 encoders and 12 decoders with 4,096 hidden state units.

\noindent\textbf{M2M-100-1.2B} It is a Transformer model released by \citet{fan2021beyond} with 24 encoders and 24 decoders with 8,192 hidden state units. 

\noindent\textbf{M2M-100-12B} It is the largest single M2M-100 model  released by \citet{fan2021beyond}, which is obtained by adding language-specific layers to M2M-100-1.2B model. 

\noindent\textbf{M2M-100-1.2B w. LG-Centric Fine-Tuning} To build a fair comparison, we also use the constructed dataset to fine-tune M2M-100-1.2B. We follow the standard fine-tuning paradigm, which uses a Transformer initialized with  M2M-100-1.2B. In this baseline, we only use LG-centric data to train models. We simply merge all translation pairs related to language LG together to get the mixed training data. Like our model does,  we also add language code in the encoder and decoder parts. 

\noindent\textbf{M2M-100-1.2B w. Multilingual Fine-Tuning} In order to establish an equitable comparison, the constructed dataset was utilized to fine-tune M2M-100-1.2B. All translation data was amalgamated for the purpose of fine-tuning M2M-100-1.2B in this baseline. Correspondingly, language codes were incorporated in both the encoder and decoder components, as is done in our model. 

\subsection{Settings and Metric}

\noindent\textbf{Training Details} The training code is bulit on the code repository  fairseq\footnote{\url{https://github.com/facebookresearch/fairseq/tree/main/examples/m2m_100}.}. Each flow is initialized with a pre-trained M2M-100-1.2B model. We train all models using Adam optimizer with $\beta_1=0.9$, $\beta_2=0.999$, the learning rate is set to 1e-4, and the max token number is set as 8,000. The training of all centric languages is conducted in random order: En, De, Ne, Az, Ceb, Ar, Zh. We split the whole dataset into 70 shards. And the whole training process takes around 15 days on 32 A100 GPUs. 

\noindent\textbf{Metric} We use the same evaluation metric (spBLEU) in the \textit{Flores-101} dataset. Before computing BLEU, we de-tokenized all data and then apply sentence piece tokenization for each language. It facilitates a more accurate assessment of model quality on the long-tail of low-resource languages.

\subsection{Results}
\label{sec:exp_res}

\noindent\textbf{Lego-MT is an efficient translation model, outperforming M2M-100-12B with only 10\% inference parameters} Table~\ref{tab:many-to-one} show experiment results on the \textit{Flores-101} devtest set. As we can see, Lego-MT is an efficient translation model that achieves large performance improvements over M2M-100-1.2B, with 7.4 spBLEU improvements on many-to-one translation and 5.8 spBLEU improvements on one-to-many translation. It even outperforms M2M-100-12B especially on one many-to-one settings, with a gain of 5.0 spBLEU. As a comparison, with the same training data, a shared model M2M-100-1.2B only obtains slight performance improvements, 4.3 spBLEU on many-to-one translation, and 2.5 spBLEU on one-to-many translation. These results demonstrate Lego-MT provides an effective solution by using fewer inference parameters to achieve higher results. 

\noindent\textbf{Compared with high-resource translation, low-resource translation benefits more from multi-way architectures.} We observe that the improvements achieved by Lego-MT are not equally contributed by different languages. As we can see from Table~\ref{tab:many-to-one}, X$\rightarrow$Ne, X$\rightarrow$Az, and X$\rightarrow$Ceb obtain more obvious improvements than X$\rightarrow$En, X$\rightarrow$Zh, and X$\rightarrow$De. On X$\rightarrow$Ne translation, Lego-MT even gets 13.6 improvements over M2M-100-1.2B. These results are consistent with previous studies about parameter interference in massive multilingual machine translation that low-resource translation usually suffers. With less parameter interference, Lego-MT gets higher low-resource translation results.

\noindent\textbf{Multilingual branches play a significant role in avoiding over-fitting.} As we can see from Table~\ref{tab:many-to-one}, only fine-tuning M2M-100-1.2B on language-centric data has serious over-fitting problems that the performance is dropped sharply, especially on low-resource settings, with a loss of 3.2 spBLEU on Ne$\rightarrow$X translation and 3.1 spBLEU on Az$\rightarrow$X translation. Like this baseline, Lego-MT also introduces language-specific parameters but does not show any performance drop. The key difference between Lego-MT with M2M-100-1.2B w. LG-Centric fine-tuning lies in that Lego-MT introduces multilingual branches as regularization, demonstrating that the unified space can avoid catastrophic forgetting.

\noindent\textbf{Lego-MT supports efficient training, which is 28.2$\times$ faster than multi-way training.} For simplification, we implement an 8-branch architecture where Lego-MT and the multi-way model both have 8 branches for encoder and decoders. We use Chinese-centric data in the first shard and select 7 Zh$\rightarrow$X and X$\rightarrow$Zh translations as a small training set, which includes high-resource languages (Be, De, Fa, Jv) and low-resource languages (Ne, Pa, Sw).
For two models, all 8 encoder branches are initialized with the encoder part of M2M-100-418M, and all 8 decoder branches are initialized with the decoder part of M2M-100-418M.  Due to the large parameter size, the multi-way model requires fewer tokens in a single batch. Lego-MT has less parameter size during each inference and thus can support more tokens in a single batch. For a fair comparison, we use the same settings for two models and set the number of tokens in a single batch to 3K. Due to the low GPU efficiency issues, the multi-way model takes 16.9 hours to finish one shard training on average while Lego-MT only takes 0.6 hours. 

\begin{table}[t!]
\centering
\footnotesize
\tabcolsep=3pt
\begin{tabular}{l|ccc}
\toprule
\textbf{Method}  & \textbf{\#Tokens}  & \textbf{Size (GB)} & \textbf{Time (Hour)}\\
\toprule
Multi-Way Training &  3,000 & 60.9 & 16.9 \\
Lego-MT Training & 3,000 & 10.3 & 0.6  \\
\bottomrule
\end{tabular}
\caption{Training efficiency of Lego-MT and a multi-way model. ``\#Tokens'' represent the maximum tokens in a single batch during training. For a fair comparison, we initialize an 8-branch Lego-MT and an 8-branch multi-way model with M2M-100-418M as initialization.  Size represents the size of the loaded parameters. ``Time'' represents the total time of completing all data (We select a small subset of training data for evaluation). In Lego-MT, we use a parallel thread for branch switching, which does not affect the running time. Lego-MT supports efficient training, which achieves 28.2$\times$ speedups over multi-way training.  }
\end{table}

\noindent\textbf{The total training cost of Lego-MT is only about twice that of M2M-1.2B fine-tuning.} In the first stage, we load a multilingual encoder-decoder and a single language-specific encoder, and in the second stage, we load a multilingual encoder-decoder and a single language-specific decoder. Compared to M2M-1.2B, the additional computations come from training language-specific parameters. Since the language-specific branch has the same size as the multilingual branch, the training costs only double. We believe that the training costs for such a model are reasonable, given its one-time training feature. In real-world applications with unlimited data, inference costs are more critical than training costs. The advantage of Lego-MT is that it largely improves translation performance without incurring additional inference costs.

\section{Analysis on Lego-MT}

\label{sec:branch_ablation}
\noindent\textbf{Ablation studies on triple-Flow training} We design three flows in Lego-MT: Mix-Flow, Enc-Flow, and Dec-Flow. Mix-FLow contains a multilingual encoder and a multilingual decoder, which is essential in regularizing language-specific training. We start from M-Flow and see how Enc-Flow and Dec-Flow affect the final performance, which gives more insights into the design of our framework. For simplification, we use Chinese-centric data in top-10 shards and select 7 Zh$\rightarrow$X and X$\rightarrow$Zh translation pairs as a small training set, which includes high-resource languages (Be, De, Fa, Jv) and low-resource languages (Ne, Pa, Sw). We train Lego-MT on the selected set and observe results in Table~\ref{tab:path}.  We can see that jointly training Enc-Flow and Mix-Flow boosts the performance in most directions. In contrast, jointly training Dec-Flow and Mix-Flow causes large performance degeneration. It is mainly because that language-specific decoder may cause a large distribution shift on multilingual encoders, resulting in catastrophic forgetting. That's why we split the training into two stages and keeps Dec-Flow in the second stage.

%%%%%%%%%%%%%%%%%%%%%%%%%%%%%%%%%
\begin{table}[!t]
% {\texttt{zh}} & ${\texttt{en}}$
\centering
\footnotesize
 \tabcolsep=2pt
\resizebox{0.98\linewidth}{!}{
\begin{tabular}{c|cc|cc|cc}
\toprule
\multirow{2}{*}{\textbf{Lang}}  
& \multicolumn{2}{c|}{\textbf{Mix-Flow }}  & \multicolumn{2}{c|}{\textbf{Dec-Flow} + \textbf{Mix-Flow }} & \multicolumn{2}{c}{\textbf{Enc-Flow} + \textbf{Mix-Flow}}\\
\cmidrule{2-7}
 & {x$\rightarrow$zh} & {zh$\rightarrow$x} & {x$\rightarrow$zh} & {zh$\rightarrow$x} & {x$\rightarrow$zh} & {zh$\rightarrow$x} \\
\midrule

Be 		& 	8.1 	& 	3.5  	& 	7.5 	& 	3.3 	& 	13.1 	& 	6.1 	\\ 
De 		& 	16.0 	& 	14.7 	& 	14.0 	& 	13.4 	& 	22.1 	& 	19.2 	\\ 
Fa 		& 	13.1 	& 	11.9 	& 	11.6 	& 	11.3 	& 	17.7 	& 	15.1 	\\ 
Jv 	& 	6.3 	& 	3.1 	& 	6.0 	& 	3.1 	& 	8.8 	& 	3.3 	\\ 
Ne & 	9.6 	& 	5.6 	& 	8.7 	& 	4.2 	& 	7.2 	& 	4.0 	\\ 
Pa 	& 	1.4 	& 	1.1  	& 	1.2 	& 	0.5 	& 	1.8 	& 	0.5 	\\ 
Sw 	& 	8.5 	& 	8.6 	& 	7.5 	& 	6.6 	& 	12.4 	& 	12.4 	\\ 
\midrule
AVG. &	9.0 &	6.9 &	8.1 &	6.1 &	11.9 &	8.7 \\ 

\bottomrule
\end{tabular}}
\caption{Ablation studies on Triple-Flow training. $\rightarrow$zh refers to the results of translating to zh and zh$\rightarrow$ refers to the results of translating from zh. Dec-Flow brings a large performance drop. }
\label{tab:path}
\end{table}
%%%%%%%%%%%%%%%%%%%%%%%%%%%%%%%%%%

%%%%%%%%%%%%%%%%%%%%%%%%%%%%%%%%%%
\begin{table}[!t]
    \centering
    \footnotesize
    \resizebox{\linewidth}{!}{
    \begin{tabular}{lccccc}
    \toprule
    \textbf{Model}  & \textbf{Ceb$\rightarrow$Ha}  & \textbf{Ceb$\rightarrow$Ig}  & \textbf{Ceb$\rightarrow$Ln} & \textbf{Ceb$\rightarrow$Yo} & \textbf{AVG.}\\
   \midrule
    M-1.2B & 5.5 & 5.9 & 0.9 & 2.4 & 3.7 \\
    M-12B & 8.7 & 10.8 & 0.9 & 2.9 & 5.8 \\
    M-FT & 6.4 & 6.6 & 0.8 & 2.1 & 4.0 \\
    Lego-MT & \textbf{12.5} & \textbf{13.9} & \textbf{2.3} & \textbf{3.2} & \textbf{8.0} \\ 
    \midrule
    \midrule
      \textbf{Model} & \textbf{Ha$\rightarrow$Ceb}  & \textbf{Ig$\rightarrow$Ceb}  & \textbf{Ln$\rightarrow$Ceb} & \textbf{Yo$\rightarrow$Ceb} & \textbf{AVG.}\\
     \midrule
    
    M-1.2B & 7.4 & 7.5 & 4.2 & 3.4 & 5.6\\
    M-12B &  8.8 & 8.8 & 3.8 & 4.1 & 6.4 \\
    M-FT & 7.5 & 8.6 & 3.7 & 4.8 & 6.2 \\
    Lego-MT & \textbf{12.3} & \textbf{12.3} & \textbf{6.2} & \textbf{6.7} & \textbf{9.4}  \\ 
    
    \midrule
    \midrule
    
    \textbf{Model}   & \textbf{X$\rightarrow$Ast} & \textbf{X$\rightarrow$Da} &\textbf{ X$\rightarrow$Hu} & \textbf{X$\rightarrow$Lo}& \textbf{AVG.} \\ 
     \midrule
        M-1.2B & \textbf{16.7} & 22.0 & 17.7 & 4.8 & 15.3 \\         
        M-12B & 13.0 & 23.3 & 19.1 & \textbf{9.0} & 16.1 \\      
        
        M-FT & 13.8 & 23.2 & 17.8 & 0.9 & 13.9 \\  
        Lego-MT & 15.4 & \textbf{25.4} & \textbf{20.1} & 5.6 &  \textbf{16.6} \\         
    \midrule
       \textbf{Model}    & \textbf{Ast$\rightarrow$X} & \textbf{Da$\rightarrow$X} & \textbf{Hu$\rightarrow$X} & \textbf{Lo$\rightarrow$X} &  \textbf{AVG.} \\ 
    \midrule
        M-1.2B & 13.2 & 18.3 & 16.0 & 6.6 & 13.5 \\         
        M-12B & 15.2 & \textbf{20.9} & \textbf{18.2} & 8.8 &  \textbf{15.8} \\
        M-FT & 14.4 & 17.9 & 15.5 & 5.8 & 13.4 \\  
        Lego-MT & \textbf{15.5} & 20.8 & 18.1 & \textbf{8.9} & \textbf{15.8} \\ 
    \bottomrule
    \end{tabular}}
    \caption{The results on unseen directions that are not covered by the constructed dataset. ``M'' means M2M-100. Lego-MT shows the best generalization results. ``M-FT'' means M2M-100-1.2B w. Multilingual FT. }
    \label{tab:unseen}
\end{table}
%%%%%%%%%%%%%%%%%%%%%%%%%%%%%%%%%%

\noindent\textbf{Analysis on inference path section} Due to the plug-and-play features, there are several possible inference paths for a single translation direction. At the inference stage, there are three alternative solutions for language-centric translation: Mix-Flow, Enc-Flow, and Dec-Flow. Figure~\ref{fig:overivew} shows the comparison between these inference paths. For low-resource languages (eg., Ceb, Az, Ne), Mix-Flow (M-encoder + M-decoder) works better than either Enc-Flow (E-encoder + M-decoder) or Dec-Flow (M-encoder + D-decoder). High-resource languages (eg., En,De,Zh, Ar) prefer language-specific branches. Dec-Flow (a multilingual encoder and a language-specific decoder) achieves better performance among these paths.  This demonstrates that specific parameters are  more important when the amount of data in a language is huge. In summary, the Mix-Flow (M-encoder + M-decoder)  is recommended for inference tasks with low-resource languages, and the Dec-FLow  (M-encoder + D-decoder) is more appropriate for high-resource languages.

\begin{figure}[!t]
    \centering
    \includegraphics[width=0.6\linewidth]{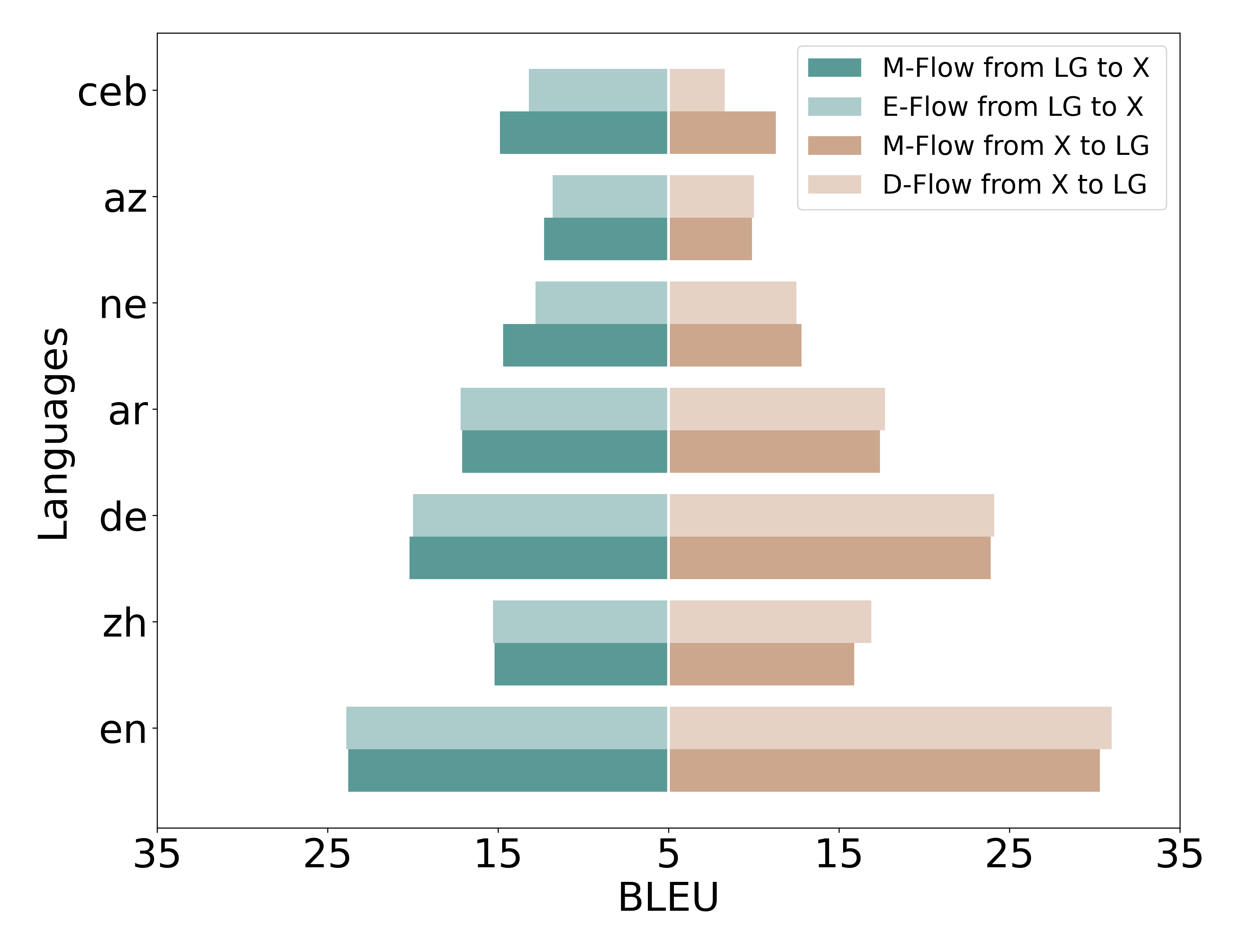}
    \caption{The comparison between different inference paths. For low-resource languages (eg. ceb, az, ne), Mix-Flow (M-encoder + M-decoder) works better than either Enc-Flow (E-encoder + M-decoder) or Dec-Flow (M-encoder + D-decoder). For high-resource languages (eg., en, de, zh, ar),  Dec-Flow (a multilingual encoder and a language-specific decoder) achieves better performance among these paths.  }
    \label{fig:unified_space}
\end{figure}

\begin{figure}[!t]
    \centering
    \includegraphics[width=0.6\linewidth]{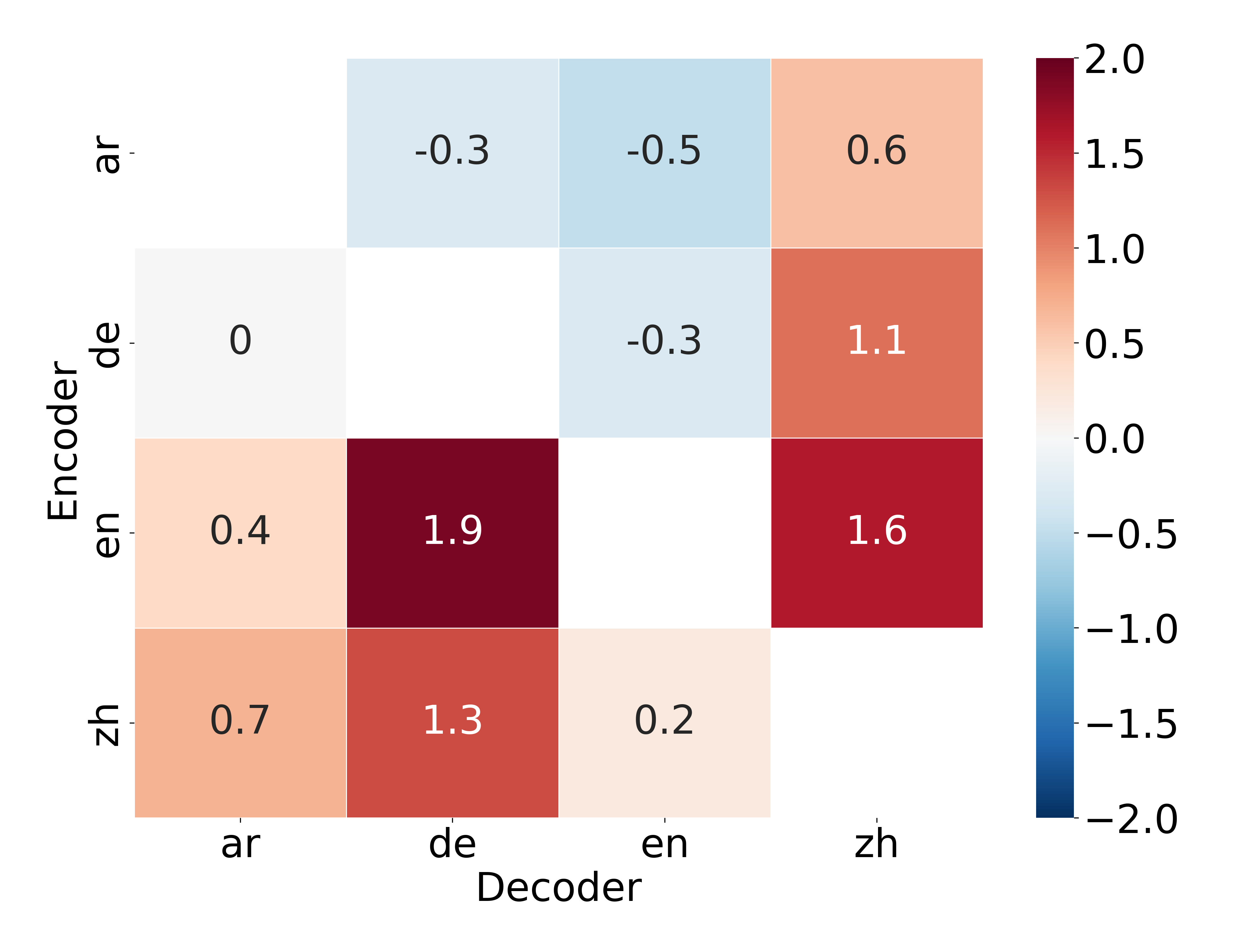}
    \caption{The spBLEU gap between Mix-Flow (a multilingual encoder and a multilingual decoder) and unseen language-specific Flow (the combination of a language-specific encoder and a language-specific decoder). Positive numbers mean the results of the language-specific Flow are better than that of the M-Flow. The unseen language-specific Flow achieves better results on 9 out of 12 directions, demonstrating that Lego-MT can learn the alignment for different branches. }
    \label{fig:unified_space1}
\end{figure}

%%%%%%%%%%%%%%%%%%%%%%%%%%%%%%%%%%
 \begin{table*}[!t]
    \centering
    \footnotesize
    \resizebox{\linewidth}{!}{
    \begin{tabular}{c|ccccccccccc|c}
    \toprule
        \textbf{Model} & \textbf{Ast$\rightarrow$X} & \textbf{Hu$\rightarrow$X} & \textbf{Da$\rightarrow$X} & \textbf{Lo$\rightarrow$X} & \textbf{En$\rightarrow$X} & \textbf{De$\rightarrow$X} & \textbf{Ar$\rightarrow$X} & \textbf{Az$\rightarrow$X} & \textbf{Ceb$\rightarrow$X} & \textbf{Ne$\rightarrow$X} & \textbf{Zh$\rightarrow$X} & \textbf{AVG.} \\
        \midrule
        Multilingual FT & 3.8 & 2.6 & 2.7 & 6.3 & 7.6 & 4.2 & 1.7 & 2.6 & 0.9 & 6.2 & 3.7 & 3.9 \\ 
        Lego-MT & \textbf{14.2} & \textbf{10.1} & \textbf{11.9} & \textbf{17.5} & \textbf{20.6} & \textbf{14.1} & \textbf{6.9} & \textbf{12.2} & \textbf{5.8} & \textbf{17.7} & \textbf{11.7} & \textbf{13.1} \\ 
        \midrule
        \textbf{Model} & \textbf{X$\rightarrow$Ast} & \textbf{X$\rightarrow$Hu} & \textbf{X$\rightarrow$Da} & \textbf{X$\rightarrow$Lo} & \textbf{X$\rightarrow$En} & \textbf{X$\rightarrow$De} & \textbf{X$\rightarrow$Ar} & \textbf{X$\rightarrow$Az} & \textbf{X$\rightarrow$Ceb} & \textbf{X$\rightarrow$Ne} & \textbf{X$\rightarrow$Zh} & \textbf{AVG.} \\ 
        \midrule
        Multilingual FT & 4.9 & 2.2 & 1.6 & 6.9 & 10.8 & 3.0 & 0.5 & 2.7 & 0.7 & 7.8 & 4.3 & 4.1 \\ 
        Lego-MT & \textbf{14.1} & \textbf{8.8} & \textbf{10.5} & \textbf{17.8} & \textbf{24.1} & \textbf{13.7} & \textbf{3.6} & \textbf{11.0} & \textbf{4.0} & \textbf{21.2} & \textbf{11.8} & \textbf{12.9} \\ 
    \bottomrule    
    \end{tabular}}
    \caption{Commencing with random initialization, $1/7$ of the data is trained and the performance differential between Multilingual-FT and Lego-MT is evaluated. The findings indicate that the mean performance of Lego-MT across all language orientations surpasses that of Multilingual-FT and exhibits a more rapid convergence rate.}
    \label{tab:random}
\end{table*}
%%%%%%%%%%%%%%%%%%%%%%%%%%%%%%%%%%

%%%%%%%%%%%%%%%%%%%%%%%%%%%%%%%%%%
\begin{table}[t!]
\centering
\footnotesize
\begin{tabular}{l|ccc}
\toprule
\textbf{Model}  & \textbf{X$\rightarrow$En}  & \textbf{En$\rightarrow$X} & \textbf{AVG.}\\
\toprule
ChatGPT zero-shot & 27.9  & 23.9 &  25.9 \\
ChatGPT eight-shot & \textbf{31.9} & 24.7 & \textbf{28.3} \\
Lego-MT & 30.2 & \textbf{25.7} &  28.0 \\
\bottomrule
\end{tabular}
\caption{Comparison of ChatGPT and Lego-MT: zero-shot and eight-shot results. While ChatGPT lags behind Lego-MT in zero-shot performance, it outperforms Lego-MT in the X$\rightarrow$En direction with eight-shot. However, in the En$\rightarrow$X direction, ChatGPT falls behind Lego-MT even with eight-shot.}
\label{tab:chatgpt_vs_lego}
\end{table}
%%%%%%%%%%%%%%%%%%%%%%%%%%%%%%%%%%

\noindent\textbf{Lego-MT can learn the align different branches into a unified space.} During training, we propose a triple-flow way to train Lego-MT. These three flows contain Mix-Flow, Dec-Flow, and Enc-Flow. To evaluate the quality of the hidden representations, we conduct experiments by directly using a language-specific encoder and a language-specific decoder for inference. Since such combinations do not occur in the training phase, it can evaluate the quality of the unified hidden space. 
We randomly combine the language-specific encoder and the language-specific decoder of four high-resource languages (En, De, Zh, Ar) with 12 translation directions. Figure~\ref{fig:unified_space1} shows the performance of directly combining language-specific encoder and decoder. We find that such unseen combinations can get better results in most translation directions (9 out of 12).  These results prove that Lego-MT can effectively map all languages into a unified space. In addition, it proves that the performance of high-resource languages still has room for improvement by using language-specific parameters.

\noindent\textbf{Lego-MT achieves promising results in unseen directions.} We also conduct experiments on unseen directions to evaluate Lego-MT’s performance in these scenarios, as demonstrated in Table~\ref{tab:unseen}. Distinguishing unseen translation directions can involve two scenarios: 1) The training data set lacks a specific translation direction. In this case, we start with the low-resource Ceb language and identify translation directions not included in our constructed data set. 2) The training data set lacks a direct translation between two languages. For instance, our training corpus may contain translations from Ast to En and from En to Es, but not a direct translation from Ast to Es. To address this, we randomly select four languages (Ast, Da, Hu, Lo) and evaluate the average performance on the \textit{Flores-101} devtest with one-to-many and many-to-one settings. According to all experimental results, Lego-MT significantly surpasses the Multilingual FT baseline and is on par with the M2M-100-12B.

\noindent\textbf{Lego-MT performance is independent of pre-trained model initialization and converges faster than existing pre-training pipelines.} To evaluate the necessity of pre-trained model initialization, we compare Lego-MT with the traditional multilingual pre-training pipeline that uses a single encoder-decoder model for all languages. We conduct experiments on a subset of our constructed corpus, which contains parallel data for 433 languages. We randomly initialize both models and train them on only 1/7 of the data, then measure their performance on \textit{Flores-101}. As shown in Table~\ref{tab:random}, our experimental results demonstrate that our Lego-MT model is independent of the pre-trained model initialization and achieves faster convergence than the traditional multilingual pre-training pipeline. Moreover, our Lego-MT model outperforms the traditional multilingual pre-training pipeline on most of the machine translation tasks, showing its superior generalization and adaptation ability.

\noindent\textbf{Lego-MT surpasses ChatGPT in the En$\rightarrow$X direction and is on par with ChatGPT in the X$\rightarrow$En direction, in terms of performance. } A comparative analysis between ChatGPT and Lego-MT, as shown in Table~\ref{tab:chatgpt_vs_lego}, reveals that in zero-shot performance, ChatGPT lags behind Lego-MT. However, in eight-shot performance, ChatGPT surpasses Lego-MT in the X$\rightarrow$En direction but falls short in the En$\rightarrow$X direction. The prompts utilized for ChatGPT are ``You are a helpful assistant that translates \{\textit{SOURCE\_LANG}\} to \{\textit{TARGET\_LANG}\}.'' for the system and ``Translate the following \{\textit{SOURCE\_LANG}\} text to \{\textit{TARGET\_LANG}\}: \{\textit{SOURCE\_TEXT}\}.'' for the user.

\section{Conclusion}
With the increasing scale of languages, using a single model to translate all directions brings new challenges in practice. This paper proposes an efficient training recipe, which results in a detachable multilingual translation model, Lego-MT. To validate the effectiveness of our algorithm, we develop a massive MNMT translation dataset, covering 433 languages. Results on \textit{Flores-101} show that Lego-MT-1.2B achieves large performance improvements over strong baselines under a fair comparison. It even outperforms the result of M2M-12B with a gain of 4 BLEU on many-to-one.

\bibliography{butterfly_model}

\begin{thebibliography}{36}
\expandafter\ifx\csname natexlab\endcsname\relax\def\natexlab#1{#1}\fi

\bibitem[{Aharoni et~al.(2019)Aharoni, Johnson, and
  Firat}]{aharoni2019massively}
Roee Aharoni, Melvin Johnson, and Orhan Firat. 2019.
\newblock Massively multilingual neural machine translation.
\newblock \emph{arXiv preprint arXiv:1903.00089}.

\bibitem[{Bapna et~al.(2019)Bapna, Arivazhagan, and Firat}]{bapna2019simple}
Ankur Bapna, Naveen Arivazhagan, and Orhan Firat. 2019.
\newblock Simple, scalable adaptation for neural machine translation.
\newblock \emph{arXiv preprint arXiv:1909.08478}.

\bibitem[{Bapna and Firat(2019)}]{bapna-firat-2019-simple}
Ankur Bapna and Orhan Firat. 2019.
\newblock \href {https://doi.org/10.18653/v1/D19-1165} {Simple, scalable
  adaptation for neural machine translation}.
\newblock In \emph{Proceedings of the 2019 Conference on Empirical Methods in
  Natural Language Processing and the 9th International Joint Conference on
  Natural Language Processing (EMNLP-IJCNLP)}, pages 1538--1548, Hong Kong,
  China. Association for Computational Linguistics.

\bibitem[{Costa{-}juss{\`{a}} et~al.(2022)Costa{-}juss{\`{a}}, Cross,
  {\c{C}}elebi, Elbayad, Heafield, Heffernan, Kalbassi, Lam, Licht, Maillard,
  Sun, Wang, Wenzek, Youngblood, Akula, Barrault, Gonzalez, Hansanti, Hoffman,
  Jarrett, Sadagopan, Rowe, Spruit, Tran, Andrews, Ayan, Bhosale, Edunov, Fan,
  Gao, Goswami, Guzm{\'{a}}n, Koehn, Mourachko, Ropers, Saleem, Schwenk, and
  Wang}]{DBLP:journals/corr/abs-2207-04672}
Marta~R. Costa{-}juss{\`{a}}, James Cross, Onur {\c{C}}elebi, Maha Elbayad,
  Kenneth Heafield, Kevin Heffernan, Elahe Kalbassi, Janice Lam, Daniel Licht,
  Jean Maillard, Anna Sun, Skyler Wang, Guillaume Wenzek, Al~Youngblood, Bapi
  Akula, Lo{\"{\i}}c Barrault, Gabriel~Mejia Gonzalez, Prangthip Hansanti, John
  Hoffman, Semarley Jarrett, Kaushik~Ram Sadagopan, Dirk Rowe, Shannon Spruit,
  Chau Tran, Pierre Andrews, Necip~Fazil Ayan, Shruti Bhosale, Sergey Edunov,
  Angela Fan, Cynthia Gao, Vedanuj Goswami, Francisco Guzm{\'{a}}n, Philipp
  Koehn, Alexandre Mourachko, Christophe Ropers, Safiyyah Saleem, Holger
  Schwenk, and Jeff Wang. 2022.
\newblock No language left behind: Scaling human-centered machine translation.
\newblock \emph{CoRR}, abs/2207.04672.

\bibitem[{Dabre et~al.(2020)Dabre, Chu, and Kunchukuttan}]{survey-mnmt}
Raj Dabre, Chenhui Chu, and Anoop Kunchukuttan. 2020.
\newblock \href {https://doi.org/10.1145/3406095} {A survey of multilingual
  neural machine translation}.
\newblock \emph{ACM Comput. Surv.}, 53(5).

\bibitem[{Dai et~al.(2022)Dai, Dong, Ma, Zheng, Sui, Chang, and
  Wei}]{DBLP:conf/acl/Dai0MZSCW22}
Damai Dai, Li~Dong, Shuming Ma, Bo~Zheng, Zhifang Sui, Baobao Chang, and Furu
  Wei. 2022.
\newblock Stablemoe: Stable routing strategy for mixture of experts.
\newblock In \emph{Proceedings of the 60th Annual Meeting of the Association
  for Computational Linguistics (Volume 1: Long Papers), {ACL} 2022, Dublin,
  Ireland, May 22-27, 2022}, pages 7085--7095. Association for Computational
  Linguistics.

\bibitem[{Dalmia et~al.(2022)Dalmia, Okhonko, Lewis, Edunov, Watanabe, Metze,
  Zettlemoyer, and Mohamed}]{dalmia2022legonn}
Siddharth Dalmia, Dmytro Okhonko, Mike Lewis, Sergey Edunov, Shinji Watanabe,
  Florian Metze, Luke Zettlemoyer, and Abdelrahman Mohamed. 2022.
\newblock \href {http://arxiv.org/abs/2206.03318} {Legonn: Building modular
  encoder-decoder models}.

\bibitem[{Dong et~al.(2015)Dong, Wu, He, Yu, and Wang}]{dong2015multi}
Daxiang Dong, Hua Wu, Wei He, Dianhai Yu, and Haifeng Wang. 2015.
\newblock Multi-task learning for multiple language translation.
\newblock In \emph{Proceedings of the Annual Meeting of the Association for
  Computational Linguistics(ACL)}.

\bibitem[{Du et~al.(2022)Du, Huang, Dai, Tong, Lepikhin, Xu, Krikun, Zhou, Yu,
  Firat et~al.}]{du2022glam}
Nan Du, Yanping Huang, Andrew~M Dai, Simon Tong, Dmitry Lepikhin, Yuanzhong Xu,
  Maxim Krikun, Yanqi Zhou, Adams~Wei Yu, Orhan Firat, et~al. 2022.
\newblock Glam: Efficient scaling of language models with mixture-of-experts.
\newblock In \emph{International Conference on Machine Learning}, pages
  5547--5569. PMLR.

\bibitem[{Fan et~al.(2021)Fan, Bhosale, Schwenk, Ma, El-Kishky, Goyal, Baines,
  Celebi, Wenzek, Chaudhary et~al.}]{fan2021beyond}
Angela Fan, Shruti Bhosale, Holger Schwenk, Zhiyi Ma, Ahmed El-Kishky,
  Siddharth Goyal, Mandeep Baines, Onur Celebi, Guillaume Wenzek, Vishrav
  Chaudhary, et~al. 2021.
\newblock Beyond english-centric multilingual machine translation.
\newblock \emph{J. Mach. Learn. Res.}, 22(107):1--48.

\bibitem[{Fedus et~al.(2021)Fedus, Zoph, and Shazeer}]{fedus2021switch}
William Fedus, Barret Zoph, and Noam Shazeer. 2021.
\newblock Switch transformers: Scaling to trillion parameter models with simple
  and efficient sparsity.

\bibitem[{Firat et~al.(2016)Firat, Cho, and Bengio}]{firat2016multi}
Orhan Firat, Kyunghyun Cho, and Yoshua Bengio. 2016.
\newblock Multi-way, multilingual neural machine translation with a shared
  attention mechanism.
\newblock In \emph{Proceedings of the Conference of the North {A}merican
  Chapter of the Association for Computational Linguistics: Human Language
  Technologies (NAACL-HLT)}.

\bibitem[{Goyal et~al.(2022)Goyal, Gao, Chaudhary, Chen, Wenzek, Ju, Krishnan,
  Ranzato, Guzman, and Fan}]{goyal2022flores}
Naman Goyal, Cynthia Gao, Vishrav Chaudhary, Peng-Jen Chen, Guillaume Wenzek,
  Da~Ju, Sanjana Krishnan, Marc’Aurelio Ranzato, Francisco Guzman, and Angela
  Fan. 2022.
\newblock The flores-101 evaluation benchmark for low-resource and multilingual
  machine translation.
\newblock \emph{Transactions of the Association for Computational Linguistics},
  10:522--538.

\bibitem[{Gu et~al.(2018)Gu, Hassan, Devlin, and Li}]{gu2018universal}
Jiatao Gu, Hany Hassan, Jacob Devlin, and Victor~O.K. Li. 2018.
\newblock Universal neural machine translation for extremely low resource
  languages.
\newblock In \emph{Proceedings of the Conference of the North {A}merican
  Chapter of the Association for Computational Linguistics: Human Language
  Technologies (NAACL-HLT)}.

\bibitem[{Ha et~al.(2016{\natexlab{a}})Ha, Niehues, and
  Waibel}]{ha-etal-2016-toward}
Thanh-Le Ha, Jan Niehues, and Alex Waibel. 2016{\natexlab{a}}.
\newblock \href {https://aclanthology.org/2016.iwslt-1.6} {Toward multilingual
  neural machine translation with universal encoder and decoder}.
\newblock In \emph{Proceedings of the 13th International Conference on Spoken
  Language Translation}, Seattle, Washington D.C. International Workshop on
  Spoken Language Translation.

\bibitem[{Ha et~al.(2016{\natexlab{b}})Ha, Niehues, and
  Waibel}]{2017-1-1-model}
Thanh-Le Ha, Jan Niehues, and Alex Waibel. 2016{\natexlab{b}}.
\newblock \href {https://aclanthology.org/2016.iwslt-1.6} {Toward multilingual
  neural machine translation with universal encoder and decoder}.
\newblock In \emph{Proceedings of the 13th International Conference on Spoken
  Language Translation}, Seattle, Washington D.C. International Workshop on
  Spoken Language Translation.

\bibitem[{Jacobs et~al.(1991)Jacobs, Jordan, Nowlan, and
  Hinton}]{jacobs1991adaptive}
Robert~A Jacobs, Michael~I Jordan, Steven~J Nowlan, and Geoffrey~E Hinton.
  1991.
\newblock Adaptive mixtures of local experts.
\newblock \emph{Neural computation}, 3(1):79--87.

\bibitem[{Ji et~al.(2020)Ji, Zhang, Duan, Zhang, Chen, and Luo}]{ji2020cross}
Baijun Ji, Zhirui Zhang, Xiangyu Duan, Min Zhang, Boxing Chen, and Weihua Luo.
  2020.
\newblock Cross-lingual pre-training based transfer for zero-shot neural
  machine translation.
\newblock In \emph{Proceedings of the AAAI conference on artificial
  intelligence}, volume~34, pages 115--122.

\bibitem[{Johnson et~al.(2017)Johnson, Schuster, Le, Krikun, Wu, Chen, Thorat,
  Vi{\'e}gas, Wattenberg, Corrado et~al.}]{johnson2017google}
Melvin Johnson, Mike Schuster, Quoc~V Le, Maxim Krikun, Yonghui Wu, Zhifeng
  Chen, Nikhil Thorat, Fernanda Vi{\'e}gas, Martin Wattenberg, Greg Corrado,
  et~al. 2017.
\newblock Google’s multilingual neural machine translation system: Enabling
  zero-shot translation.
\newblock \emph{Transactions of the Association for Computational Linguistics},
  5:339--351.

\bibitem[{Lee et~al.(2017)Lee, Cho, and Hofmann}]{lee2017fully}
Jason Lee, Kyunghyun Cho, and Thomas Hofmann. 2017.
\newblock Fully character-level neural machine translation without explicit
  segmentation.
\newblock \emph{Transactions of the Association for Computational Linguistics},
  5:365--378.

\bibitem[{Lepikhin et~al.(2020)Lepikhin, Lee, Xu, Chen, Firat, Huang, Krikun,
  Shazeer, and Chen}]{lepikhin2020gshard}
Dmitry Lepikhin, HyoukJoong Lee, Yuanzhong Xu, Dehao Chen, Orhan Firat, Yanping
  Huang, Maxim Krikun, Noam Shazeer, and Zhifeng Chen. 2020.
\newblock Gshard: Scaling giant models with conditional computation and
  automatic sharding.
\newblock \emph{arXiv preprint arXiv:2006.16668}.

\bibitem[{Li and Gong(2021)}]{li2021robust}
Xian Li and Hongyu Gong. 2021.
\newblock Robust optimization for multilingual translation with imbalanced
  data.
\newblock \emph{Advances in Neural Information Processing Systems (NeurIPS)}.

\bibitem[{Lin et~al.(2020)Lin, Pan, Wang, Qiu, Feng, Zhou, and Li}]{lin2020pre}
Zehui Lin, Xiao Pan, Mingxuan Wang, Xipeng Qiu, Jiangtao Feng, Hao Zhou, and
  Lei Li. 2020.
\newblock \href {http://arxiv.org/abs/https://arxiv.org/abs/2010.03142}
  {Pre-training multilingual neural machine translation by leveraging alignment
  information}.
\newblock In \emph{the Conference on Empirical Methods in Natural Language
  Processing (EMNLP)}.

\bibitem[{Lin et~al.(2021)Lin, Wu, Wang, and Li}]{lin2021learning}
Zehui Lin, Liwei Wu, Mingxuan Wang, and Lei Li. 2021.
\newblock \href {http://arxiv.org/abs/https://arxiv.org/abs/2105.09259}
  {Learning language specific sub-network for multilingual machine
  translation}.
\newblock In \emph{the 59th Annual Meeting of the Association for Computational
  Linguistics (ACL)}.

\bibitem[{Liu et~al.(2020)Liu, Gu, Goyal, Li, Edunov, Ghazvininejad, Lewis, and
  Zettlemoyer}]{liu2020multilingual}
Yinhan Liu, Jiatao Gu, Naman Goyal, Xian Li, Sergey Edunov, Marjan
  Ghazvininejad, Mike Lewis, and Luke Zettlemoyer. 2020.
\newblock Multilingual denoising pre-training for neural machine translation.
\newblock \emph{Transactions of the Association for Computational Linguistics},
  8:726--742.

\bibitem[{Lyu et~al.(2020)Lyu, Son, Yang, and Bae}]{lyu-etal-2020-revisiting}
Sungwon Lyu, Bokyung Son, Kichang Yang, and Jaekyoung Bae. 2020.
\newblock \href {https://doi.org/10.18653/v1/2020.emnlp-main.476} {Revisiting
  {M}odularized {M}ultilingual {NMT} to {M}eet {I}ndustrial {D}emands}.
\newblock In \emph{Proceedings of the 2020 Conference on Empirical Methods in
  Natural Language Processing (EMNLP)}, pages 5905--5918, Online. Association
  for Computational Linguistics.

\bibitem[{Neubig and Hu(2018)}]{neubig2018rapid}
Graham Neubig and Junjie Hu. 2018.
\newblock Rapid adaptation of neural machine translation to new languages.
\newblock In \emph{Proceedings of the Conference on Empirical Methods in
  Natural Language Processing (EMNLP)}.

\bibitem[{Pan et~al.(2021)Pan, Wu, Wang, and Li}]{pan2021contrastive}
Xiao Pan, Liwei Wu, Mingxuan Wang, and Lei Li. 2021.
\newblock \href {http://arxiv.org/abs/https://arxiv.org/abs/2105.09501}
  {Contrastive learning for many-to-many multilingual neural machine
  translation}.
\newblock In \emph{the 59th Annual Meeting of the Association for Computational
  Linguistics (ACL)}.

\bibitem[{Sachan and Neubig(2018)}]{sachan-neubig-2018-parameter}
Devendra Sachan and Graham Neubig. 2018.
\newblock \href {https://doi.org/10.18653/v1/W18-6327} {Parameter sharing
  methods for multilingual self-attentional translation models}.
\newblock In \emph{Proceedings of the Third Conference on Machine Translation:
  Research Papers}, pages 261--271, Brussels, Belgium. Association for
  Computational Linguistics.

\bibitem[{Shazeer et~al.(2017)Shazeer, Mirhoseini, Maziarz, Davis, Le, Hinton,
  and Dean}]{shazeer2017outrageously}
Noam Shazeer, Azalia Mirhoseini, Krzysztof Maziarz, Andy Davis, Quoc Le,
  Geoffrey Hinton, and Jeff Dean. 2017.
\newblock Outrageously large neural networks: The sparsely-gated
  mixture-of-experts layer.
\newblock \emph{arXiv preprint arXiv:1701.06538}.

\bibitem[{Song et~al.(2021)Song, Zhou, Qian, Xu, Cheng, Wang, and
  Li}]{song2021switch}
Zhenqiao Song, Hao Zhou, Lihua Qian, Jingjing Xu, Shanbo Cheng, Mingxuan Wang,
  and Lei Li. 2021.
\newblock switch-glat: Multilingual parallel machine translation via
  code-switch decoder.
\newblock In \emph{International Conference on Learning Representations
  (ICLR)}.

\bibitem[{Sun et~al.(2021)Sun, Wang, and Li}]{sun2021multilingual}
Zewei Sun, Mingxuan Wang, and Lei Li. 2021.
\newblock \href {http://arxiv.org/abs/https://arxiv.org/abs/2109.05256}
  {Multilingual translation via grafting pre-trained language models}.
\newblock In \emph{the Conference on Empirical Methods in Natural Language
  Processing (EMNLP) - Findings}.

\bibitem[{Tiedemann(2012)}]{tiedemann2012parallel}
J{\"o}rg Tiedemann. 2012.
\newblock Parallel data, tools and interfaces in {OPUS}.
\newblock In \emph{Proceedings of the Eighth International Conference on
  Language Resources and Evaluation (LREC)}.

\bibitem[{Zhang et~al.(2020)Zhang, Williams, Titov, and
  Sennrich}]{zhang2020improving}
Biao Zhang, Philip Williams, Ivan Titov, and Rico Sennrich. 2020.
\newblock Improving massively multilingual neural machine translation and
  zero-shot translation.
\newblock \emph{arXiv preprint arXiv:2004.11867}.

\bibitem[{Zhu et~al.(2021)Zhu, Feng, Zhao, Wang, and
  Li}]{DBLP:conf/emnlp/ZhuFZWL21}
Yaoming Zhu, Jiangtao Feng, Chengqi Zhao, Mingxuan Wang, and Lei Li. 2021.
\newblock Counter-interference adapter for multilingual machine translation.
\newblock In \emph{Findings of the Association for Computational Linguistics:
  {EMNLP} 2021, Virtual Event / Punta Cana, Dominican Republic, 16-20 November,
  2021}, pages 2812--2823.

\bibitem[{Zoph and Knight(2016)}]{zoph2016multi}
Barret Zoph and Kevin Knight. 2016.
\newblock Multi-source neural translation.
\newblock \emph{arXiv preprint arXiv:1601.00710}.

\end{thebibliography}
\bibliographystyle{acl_natbib}

\newpage

\appendix

\section*{Limitation}

Despite promising results, we also notice several limitations in this paper. First, we find that low-resource translation is not boosted by language-specific decoders and language-specific encoders, which require more exploration of the trade-off between parameter sharing and parameter tension. Second, the evaluation of few-shot languages still remains a large problem. Although the final training dataset covers 433 languages, we only evaluate the translation performance on the available evaluation set that only covers 86 languages since baselines do not support so many languages. More standard benchmarks are required for evaluation.

\section{Dataset construction}
\label{sec:construction}
In this section, we will describe the construction details of the Many-to-Many dataset. As shown in Table~\ref{fig:data-pipeline}, the  pipeline  mainly consists of six steps:

\begin{figure}[!h]
    \centering
    \includegraphics[trim={0cm 3.2cm 20cm 2cm},clip,scale=0.40]{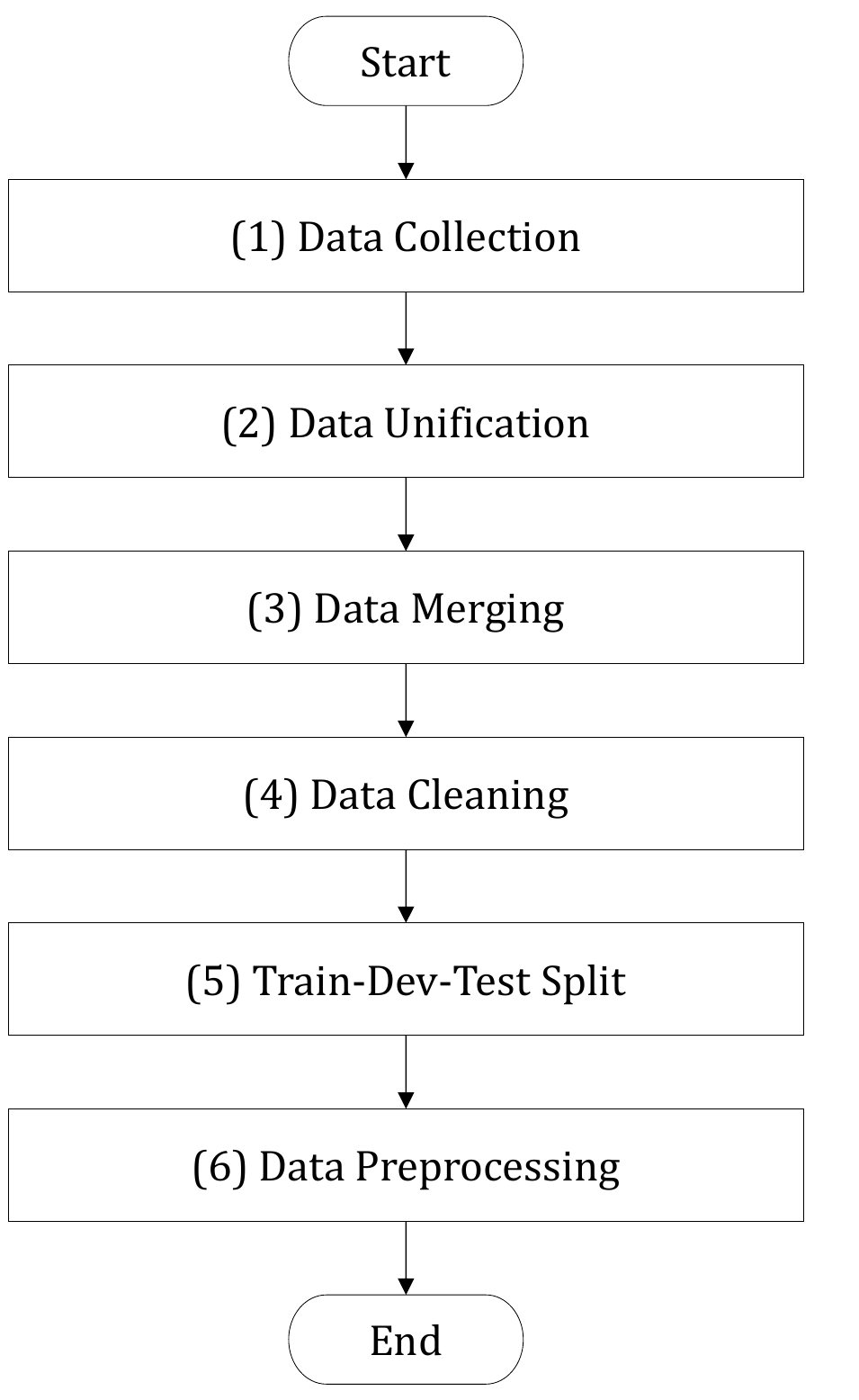}
    \caption{The construction pipeline for Many-to-Many dataset.}
    \label{fig:data-pipeline}
\end{figure}

\paragraph{Step 1: Data Collection} The raw data is collected from OPUS$\footnote{\url{https://opus.nlpl.eu/}}$, which is an open corpus that collects numerous parallel sentences from the web and overs a large number of domains from legislative to religious texts. 

\paragraph{Step 2: Data Unification} Since the OPUS includes datasets from different sources, it leads to the following two significant issues.

\textit{1) Different Language Code:} Some language in OPUS has several corresponding language codes. One of the reasons is that different corpora use different standards for language code, including ISO 639-1, ISO 639-2, ISO 639-3 or self-defined language codes in OPUS. Another reason is that some corpora append region ids at the end of language codes to distinguish the same language used in different regions. To unify language codes, we replace ISO 639-2 and ISO 639-3 language codes with ISO 639-1 language codes if the codes from ISO 639-1, ISO 639-2 and ISO 639-3 have the same language name in the code set published by SIL International (formerly known as the Summer Institute of Linguistics)$\footnote{\url{https://iso639-3.sil.org/sites/iso639-3/files/downloads/iso-639-3.tab}}$.

\textit{2) Inconsistent Operation:} Some datasets pre-tokenize their sentences in OPUS, especially for Chinese and Japanese.

Therefore, we remove the region id if the language code ends with a region id. All replaced language codes are shown in Table~\ref{tab:replacement}. For the language codes out of ISO 639 series, we list them and the corpus they come from in Table~\ref{tab:unknown ids}. Furthermore, we report all used language codes and the full names of their corresponding languages in our dataset in Table~\ref{tab:lang list}. Then we detokenize all sentences by removing white space and unifying our texts.

\paragraph{Step 3: Data Merging} After data unification, the parallel data is merged with the same language code pair from a different corpus.

\paragraph{Step 4: Data Cleaning} The OPUS corpus collected from the web contains some poor-quality data. The main problems are:

\textit{1) Duplication:} we use the deduplication script from fairseq$\footnote{\url{https://github.com/facebookresearch/fairseq/edit/main/examples/backtranslation/deduplicate_lines.py}}$ to remove all duplicated sentence pairs for each language pair.

\textit{2) Missing Translation:} We remove the sentence without corresponding translation or repeating itself as translation.

\textit{3) Length Mismatching:} After segmentation the sentences with white space for most languages or individual characters for Chinese and Japanese, we apply a filtering script from Moses decoder$\footnote{\url{https://github.com/moses-smt/mosesdecoder}}$ to remove the sentences that the length is more than 250 words or three times difference in length between source and target sentences.

\paragraph{Step 5: Train-Dev-Test Split} Different train-dev-test split schemes are developed based on the data quantity.  

\textit{1) A parallel data with more than 6,000 sentence pairs.} We randomly sample separately about 2,000 sentence pairs as validation and test set, respectively. And the rest is train set. 

\textit{2) A parallel data with fewer than 6,000 sentence pairs. } We take 80\%, 10\%, 10\% of all samples as train, validation, and test. 

To avoid any overlap between our training data and used benchmark test data, we filter all sentences that exist in the common benchmarks (WMT, \textit{Flores-101}) from our train and validation set.

\paragraph{Step 6: Data Preprocessing} The data preprocessing consists of two main steps: 

\textit{1) Sampling:} Because the full dataset is huge, we sample some data for our training. The final dataset contains 1,307,143,514 sentence pairs, 433 languages, and 1,922 training pairs. %We evenly split high-resource language pairs into 1000 parts and copy low-resource language pairs to each part. In the final experiment setting, we only take 70 of 1000 parts as our training corpus. Table~\ref{tab:statistic} shows the number of parallel sentences for 7 core languages En, Zh, De, Ar, Ne, Az, Ceb. 

\textit{2) Preprocessing: } The data is preprocess using the SentencePiece tokenizer provided by ~\citet{fan2021beyond} with a shared vocabulary of size 128,112.

\label{sec:appendix}

% \section{Dataset Statistics}
% Our dataset contains 433 languages. Table~\ref{tab:statistic} shows the number of parallel sentences in the training set for each language.

\begin{table}[!hb]
\footnotesize
  \centering
  \resizebox{\linewidth}{!}{
    \begin{tabular}{ll|ll|ll}
    \toprule
    \multicolumn{1}{l}{\textbf{Original}} & \multicolumn{1}{l}{\textbf{Replaced}} & \multicolumn{1}{l}{\textbf{Original}} & \multicolumn{1}{l}{\textbf{Replaced}} & \multicolumn{1}{l}{\textbf{Original}} & \multicolumn{1}{l}{\textbf{Replaced}} \\
    \midrule
    ak    & aka   & es    & es\_HN & pt    & pt\_BR \\
    am    & amh   & es    & es\_EC & pt    & pt\_br \\
    ar    & ara   & es    & es\_CO & pt    & pt\_PT \\
    ar    & ar\_SY & fa    & fa\_IR & rn    & run \\
    ar    & ar\_TN & fa    & fa\_AF & rw    & kin \\
    ay    & aym   & ff    & ful   & sn    & sna \\
    az    & az\_IR & fr    & fr\_FR & so    & som \\
    bg    & bg\_BG & fr    & fr\_CA & sr    & srp \\
    bm    & bam   & fr    & fr\_BE & sr    & sr\_ME \\
    bn    & bn\_IN & fr    & fr\_ca & st    & sot \\
    ca    & cat   & ha    & hau   & sw    & swa \\
    da    & da\_DK & hi    & hi\_IN & ta    & ta\_LK \\
    de    & de\_CH & ig    & ibo   & tg    & tg\_TJ \\
    de    & de\_AT & it    & it\_IT & ti    & tir \\
    de    & de\_DE & jp    & jap   & tl    & tl\_PH \\
    es    & es\_CL & kr    & kau   & tr    & tr\_TR \\
    es    & es\_SV & kv    & kpv   & ur    & ur\_PK \\
    es    & es\_NI & ln    & lin   & vi    & vi\_VN \\
    es    & es\_UY & mg    & mlg   & wo    & wol \\
    es    & es\_PE & ms    & ms\_MY & xh    & xho \\
    es    & es\_VE & nb    & nb\_NO & yo    & yor \\
    es    & es\_AR & nds   & nds\_nl & ze    & ze\_zh \\
    es    & es\_MX & nl    & nl\_NL & ze    & ze\_en \\
    es    & es\_PA & nl    & nl\_BE & zh    & zh\_cn \\
    es    & es\_CR & nn    & nn\_NO & zh    & zh\_CN \\
    es    & es\_PR & no    & no\_nb & zhtrad & zh\_HK \\
    es    & es\_ES & ny    & nya   & zhtrad & zh\_TW \\
    es    & es\_GT & om    & orm   & zhtrad & zh\_tw \\
    es    & es\_DO & pa    & pan   & zu    & zul \\
    \bottomrule
    \end{tabular}}%
      \caption{Code Replacement List. We use the codes in the column ``Original'' to replace the codes in the column ``replaced'' if these replaced codes exist in OPUS.}

  \label{tab:replacement}%
\end{table}%

\begin{table*}[!t]
  \centering
  \footnotesize
    \begin{tabular}{c|ccccccc}
        \toprule
    \textbf{Code}  & \textbf{En} & \textbf{De} & \textbf{Ar} & \textbf{Zh} & \textbf{Ne} & \textbf{Az} & \textbf{Ceb} \\
    \midrule
    \#Sentence Pairs & 811,238,712 & 360,369,144 & 152,457,830 & 92,763,445 & 6,654,270 & 4,208,025 & 1,683,531 \\
    
    \bottomrule
    \end{tabular}%
  \caption{The number of sentence pairs for each core language in Lego-MT training, }
  \label{tab:statistic}%
\end{table*}%

\begin{table*}[!t]
\footnotesize
  \centering
    \resizebox{0.96\linewidth}{!}{
    \begin{tabular}{ll|ll|ll|ll|ll}
    \toprule
    % \textbf{code}  & \textbf{dataset} & \textbf{code}  & \textbf{dataset} \\
    \textbf{Code} & \textbf{Dataset} & \textbf{Code} &\textbf{Dataset} & \textbf{Code} & \textbf{Dataset} & \textbf{Code} &\textbf{Dataset} & \textbf{Code} &\textbf{Dataset}  \\
    \midrule

        crp    &  bible-uedin  &     cb     &  MultiCCAligned  &     sz     &  MultiCCAligned  &  sgn    &  QED  &  cycl   &  Tatoeba  \\
    tc     &  EUbookshop  &     cx     &  MultiCCAligned  &  zz     &  MultiCCAligned  &  iro    &  QED  &  nah    &  Tatoeba  \\
    zhs    &  GlobalVoices  &     ns     &  MultiCCAligned  &  ze     &  OpenSubtitles  &  mo     &  QED,Ubuntu 		\\
    zht    &  GlobalVoices  &     qd     &  MultiCCAligned  &  bh     &  QED  &  ber    &  QED,Ubuntu 		\\
    tmp    &  GNOME  &     qa     &  MultiCCAligned  &  bnt    &  QED  &  toki   &  Tatoeba 		\\
    gr     &  GNOME  &     tz     &  MultiCCAligned  &  ry     &  QED  &  kzj    &  Tatoeba 		\\
    \bottomrule
    \end{tabular}}%
      \caption{Unkown Language Codes, which are out of ISO 639 series. We can't confirm their full names.}

  \label{tab:unknown ids}%
\end{table*}%

% \begin{longtable}[!hb]
\begin{table*}[!t]
\centering
%\setlength{\tabcolsep}{4pt}
% \scriptsize
\footnotesize
  \centering
  \resizebox{\textwidth}{!}{
    \begin{tabular}{ll|ll|ll|ll|ll|rr}
    \toprule
    \multicolumn{1}{p{1mm}}{\textbf{Language}} & \multicolumn{1}{p{1mm}}{\textbf{Code}} & \multicolumn{1}{p{1mm}}{\textbf{Language}} & \multicolumn{1}{p{1mm}}{\textbf{Code}} & \multicolumn{1}{p{1mm}}{\textbf{Language}} & \multicolumn{1}{p{1mm}}{\textbf{Code}} & \multicolumn{1}{p{1mm}}{\textbf{Language}} & \multicolumn{1}{p{1mm}}{\textbf{Code}} & \multicolumn{1}{p{1mm}}{\textbf{Language}} & \multicolumn{1}{p{1mm}}{\textbf{Code}} & \multicolumn{1}{p{1mm}}{\textbf{Language}} & \multicolumn{1}{p{1mm}}{\textbf{Code}} \\
    \midrule
    Abkhazian & ab    & Corsican & co    & Iban  & iba   & Lower Sorbian & dsb   & Ossetian & os    & \multicolumn{1}{l}{Swahili (macrolanguage)} & \multicolumn{1}{l}{sw} \\
    Achinese & ace   & Cree  & cr    & Icelandic & is    & Lukpa & dop   & Ottoman Turkish (1500-1928) & ota   & \multicolumn{1}{l}{Swati} & \multicolumn{1}{l}{ss} \\
    Achuar-Shiwiar & acu   & Creek & mus   & Ido   & io    & Luo (Kenya and Tanzania) & luo   & Paite Chin & pck   & \multicolumn{1}{l}{Swedish} & \multicolumn{1}{l}{sv} \\
    Adyghe & ady   & Crimean Tatar & crh   & Igbo  & ig    & Lushootseed & lut   & Palauan & pau   & \multicolumn{1}{l}{Swiss German} & \multicolumn{1}{l}{gsw} \\
    Afar  & aa    & Croatian & hr    & Iloko & ilo   & Luxembourgish & lb    & Pali  & pi    & \multicolumn{1}{l}{Syriac} & \multicolumn{1}{l}{syr} \\
    Afrihili & afh   & Cusco Quechua & quz   & Indonesian & id    & Luyia & luy   & Pampanga & pam   & \multicolumn{1}{l}{Tachawit} & \multicolumn{1}{l}{shy} \\
    Afrikaans & af    & Czech & cs    & Ingrian & izh   & Macedonian & mk    & Pangasinan & pag   & \multicolumn{1}{l}{Tachelhit} & \multicolumn{1}{l}{shi} \\
    Aguaruna & agr   & Danish & da    & Ingush & inh   & Macedo-Romanian & rup   & Panjabi & pa    & \multicolumn{1}{l}{Tagal Murut} & \multicolumn{1}{l}{mvv} \\
    Ainu (Japan) & ain   & Dari  & prs   & Interlingua & ia    & Madurese & mad   & Papiamento & pap   & \multicolumn{1}{l}{Tagalog} & \multicolumn{1}{l}{tl} \\
    Akan  & ak    & Dinka & din   & Interlingue & ie    & Maithili & mai   & Papuan Malay & pmy   & \multicolumn{1}{l}{Tahaggart Tamahaq} & \multicolumn{1}{l}{thv} \\
    Akawaio & ake   & Drents & drt   & Inuktitut & iu    & Malagasy & mg    & Pedi  & nso   & \multicolumn{1}{l}{Tahitian} & \multicolumn{1}{l}{ty} \\
    Aklanon & akl   & Dungan & dng   & Inupiaq & ik    & Malay (individual language) & zlm   & Pennsylvania German & pdc   & \multicolumn{1}{l}{Tajik} & \multicolumn{1}{l}{tg} \\
    Albanian & sq    & Dutch & nl    & Iranian Persian & pes   & Malay (macrolanguage) & ms    & Persian & fa    & \multicolumn{1}{l}{Talossan} & \multicolumn{1}{l}{tzl} \\
    Algerian Arabic & arq   & Dutton World Speedwords & dws   & Irish & ga    & Malayalam & ml    & Phoenician & phn   & \multicolumn{1}{l}{Talysh} & \multicolumn{1}{l}{tly} \\
    American Sign Language & ase   & Dzongkha & dz    & Italian & it    & Maltese & mt    & Picard & pcd   & \multicolumn{1}{l}{Tamashek} & \multicolumn{1}{l}{tmh} \\
    Amharic & am    & Eastern Canadian Inuktitut & ike   & Jakun & jak   & Mam   & mam   & Piemontese & pms   & \multicolumn{1}{l}{Tamil} & \multicolumn{1}{l}{ta} \\
    Ancient Greek (to 1453) & grc   & Eastern Mari & mhr   & Jamaican Creole English & jam   & Mambae & mgm   & Pipil & ppl   & \multicolumn{1}{l}{Tarifit} & \multicolumn{1}{l}{rif} \\
    Ancient Hebrew & hbo   & Eastern Maroon Creole & djk   & Japanese & ja    & Mandarin Chinese & cmn   & Plateau Malagasy & plt   & \multicolumn{1}{l}{Tase Naga} & \multicolumn{1}{l}{nst} \\
    Arabic & ar    & Efik  & efi   & Javanese & jv    & Manx  & gv    & Polish & pl    & \multicolumn{1}{l}{Tatar} & \multicolumn{1}{l}{tt} \\
    Aragonese & an    & Egyptian Arabic & arz   & Jewish Babylonian Aramaic & tmr   & Maori & mi    & Portuguese & pt    & \multicolumn{1}{l}{Telugu} & \multicolumn{1}{l}{te} \\
    Armenian & hy    & Emilian & egl   & Kabyle & kab   & Marathi & mr    & Potawatomi & pot   & \multicolumn{1}{l}{Tena Lowland Quichua} & \multicolumn{1}{l}{quw} \\
    Arpitan & frp   & English & en    & Kadazan Dusun & dtp   & Marshallese & mh    & Prussian & prg   & \multicolumn{1}{l}{Tetelcingo Nahuatl} & \multicolumn{1}{l}{nhg} \\
    Asháninka & cni   & Erzya & myv   & Kalaallisut & kl    & Mesopotamian Arabic & acm   & Pushto & ps    & \multicolumn{1}{l}{Tetum} & \multicolumn{1}{l}{tet} \\
    Assamese & as    & Esperanto & eo    & Kalmyk & xal   & Miahuatlán Zapotec & zam   & Quechua & qu & \multicolumn{1}{l}{Thai} & \multicolumn{1}{l}{th} \\
    Asturian & ast   & Estonian & et    & Kamba (Kenya) & kam   & Middle English (1100-1500) & enm   & Quenya & qya   & \multicolumn{1}{l}{Tibetan} & \multicolumn{1}{l}{bo} \\
    Avaric & av    & Evenki & evn   & Kannada & kn    & Middle French (ca. 1400-1600) & frm   & Quiotepec Chinantec & chq   & \multicolumn{1}{l}{Tigrinya} & \multicolumn{1}{l}{ti} \\
    Avestan & ae    & Ewe   & ee    & Kanuri & kr    & Mikasuki & mik   & Rapanui & rap   & \multicolumn{1}{l}{Tohono O'odham} & \multicolumn{1}{l}{ood} \\
    Awadhi & awa   & Extremaduran & ext   & Kaqchikel & cak   & Mi'kmaq & mic   & Romanian & ro    & \multicolumn{1}{l}{Tok Pisin} & \multicolumn{1}{l}{tpi} \\
    Aymara & ay    & Faroese & fo    & Karelian & krl   & Min Dong Chinese & cdo   & Romansh & rm    & \multicolumn{1}{l}{Tonga (Tonga Islands)} & \multicolumn{1}{l}{to} \\
    Azerbaijani & az    & Fiji Hindi & hif   & Kashmiri & ks    & Min Nan Chinese & nan   & Romany & rom   & \multicolumn{1}{l}{Traditional Chinese} & \multicolumn{1}{l}{zhtrad} \\
    Baluchi & bal   & Fijian & fj    & Kashubian & csb   & Minangkabau & min   & Rundi & rn    & \multicolumn{1}{l}{Tsonga} & \multicolumn{1}{l}{ts} \\
    Bambara & bm    & Filipino & fil   & Kazakh & kk    & Mingrelian & xmf   & Russian & ru    & \multicolumn{1}{l}{Tswana} & \multicolumn{1}{l}{tn} \\
    Banjar & bjn   & Finnish & fi    & Kekchí & kek   & Mirandese & mwl   & Rusyn & rue   & \multicolumn{1}{l}{Tupí} & \multicolumn{1}{l}{tpw} \\
    Barasana-Eduria & bsn   & French & fr    & Khakas & kjh   & Mískito & miq   & Samoan & sm    & \multicolumn{1}{l}{Turkish} & \multicolumn{1}{l}{tr} \\
    Bashkir & ba    & Friulian & fur   & Khasi & kha   & Modern Greek (1453-) & el    & Samogitian & sgs   & \multicolumn{1}{l}{Turkmen} & \multicolumn{1}{l}{tk} \\
    Basque & eu    & Fulah & ff    & Khmer & km    & Mohawk & moh   & Sango & sg    & \multicolumn{1}{l}{Tuvalu} & \multicolumn{1}{l}{tvl} \\
    Bavarian & bar   & Galela & gbi   & K'iche' & quc   & Mongolian & mn    & Sanskrit & sa    & \multicolumn{1}{l}{Twi} & \multicolumn{1}{l}{tw} \\
    Baybayanon & bvy   & Galician & gl    & Kikuyu & kik   & Morisyen & mfe   & Santali & sat   & \multicolumn{1}{l}{Uab Meto} & \multicolumn{1}{l}{aoz} \\
    Belarusian & be    & Gan Chinese & gan   & Kinyarwanda & rw    & Moroccan Arabic & ary   & Sardinian & sc    & \multicolumn{1}{l}{Udmurt} & \multicolumn{1}{l}{udm} \\
    Bemba (Zambia) & bem   & Ganda & lg    & Kirghiz & ky    & Mossi & mos   & Saterfriesisch & stq   & \multicolumn{1}{l}{Uighur} & \multicolumn{1}{l}{ug} \\
    Bengali & bn    & Garhwali & gbm   & Klingon & tlh   & Nauru & na    & Scots & sco   & \multicolumn{1}{l}{Ukrainian} & \multicolumn{1}{l}{uk} \\
    Berom & bom   & Georgian & ka    & Koasati & cku   & Navajo & nv    & Scottish Gaelic & gd    & \multicolumn{1}{l}{Uma} & \multicolumn{1}{l}{ppk} \\
    Bhojpuri & bho   & German & de    & Kölsch & ksh   & Neapolitan & nap   & Sediq & trv   & \multicolumn{1}{l}{Umbundu} & \multicolumn{1}{l}{umb} \\
    Bislama & bi    & Gheg Albanian & aln   & Komi  & kv    & Nepali (individual language) & npi   & Serbian & sr    & \multicolumn{1}{l}{Upper Sorbian} & \multicolumn{1}{l}{hsb} \\
    Bodo (India) & brx   & Gilbertese & gil   & Komi-Permyak & koi   & Nepali (macrolanguage) & ne    & Serbo-Croatian & sh & \multicolumn{1}{l}{Urdu} & \multicolumn{1}{l}{ur} \\
    Bosnian & bs    & Goan Konkani & gom   & Kongo & kg    & Nigerian Fulfulde & fuv   & Shan  & shn   & \multicolumn{1}{l}{Uspanteco} & \multicolumn{1}{l}{usp} \\
    Breton & br    & Gothic & got   & Korean & ko    & Niuean & niu   & Shona & sn    & \multicolumn{1}{l}{Uzbek} & \multicolumn{1}{l}{uz} \\
    Brithenig & bzt   & Gronings & gos   & Kotava & avk   & Nogai & nog   & Shuar & jiv   & \multicolumn{1}{l}{Venda} & \multicolumn{1}{l}{ve} \\
    Buginese & bug   & Guadeloupean Creole French & gcf   & Kriang & ngt   & North Levantine Arabic & apc   & Shuswap & shs   & \multicolumn{1}{l}{Venetian} & \multicolumn{1}{l}{vec} \\
    Bulgarian & bg    & Guarani & gn    & Kuanyama & kj    & North Moluccan Malay & max   & Sicilian & scn   & \multicolumn{1}{l}{Vietnamese} & \multicolumn{1}{l}{vi} \\
    Buriat & bua   & Guerrero Amuzgo & amu   & Kurdish & ku    & Northern Frisian & frr   & Silesian & szl   & \multicolumn{1}{l}{Vlaams} & \multicolumn{1}{l}{vls} \\
    Burmese & my    & Guerrero Nahuatl & ngu   & Kven Finnish & fkv   & Northern Kurdish & kmr   & Sindarin & sjn   & \multicolumn{1}{l}{Volapük} & \multicolumn{1}{l}{vo} \\
    Cabécar & cjp   & Gujarati & gu    & Láadan & ldn   & Northern Sami & se    & Sindhi & sd    & \multicolumn{1}{l}{Walloon} & \multicolumn{1}{l}{wa} \\
    Camsá & kbh   & Gulf Arabic & afb   & Ladin & lld   & Northwestern Ojibwa & ojb   & Sinhala & si    & \multicolumn{1}{l}{Walser} & \multicolumn{1}{l}{wae} \\
    Catalan & ca    & Haida & hai   & Ladino & lad   & Norwegian & no    & Slovak & sk    & \multicolumn{1}{l}{Waray (Philippines)} & \multicolumn{1}{l}{war} \\
    Cebuano & ceb   & Haitian & ht    & Lakota & lkt   & Norwegian Bokmål & nb    & Slovenian & sl    & \multicolumn{1}{l}{Welsh} & \multicolumn{1}{l}{cy} \\
    Central Huasteca Nahuatl & nch   & Hakha Chin & cnh   & Lao   & lo    & Norwegian Nynorsk & nn    & Somali & so    & \multicolumn{1}{l}{Western Frisian} & \multicolumn{1}{l}{fy} \\
    Central Kurdish & ckb   & Hakka Chinese & hak   & Latgalian & ltg   & Novial & nov   & South Azerbaijani & azb   & \multicolumn{1}{l}{Western Panjabi} & \multicolumn{1}{l}{pnb} \\
    Central Sama & sml   & Hausa & ha    & Latin & la    & Nuer  & nus   & South Ndebele & nr    & \multicolumn{1}{l}{Wolaytta} & \multicolumn{1}{l}{wal} \\
    Chamorro & ch    & Hawaiian & haw   & Latvian & lv    & Nyanja & ny    & Southern Kurdish & sdh   & \multicolumn{1}{l}{Wolof} & \multicolumn{1}{l}{wo} \\
    Chavacano & cbk   & Hebrew & he    & Ligurian & lij   & Occitan (post 1500) & oc    & Southern Sami & sma   & \multicolumn{1}{l}{Wu Chinese} & \multicolumn{1}{l}{wuu} \\
    Chechen & ce    & Hiligaynon & hil   & Limburgan & li    & Old English (ca. 450-1100) & ang   & Southern Sotho & st    & \multicolumn{1}{l}{Xhosa} & \multicolumn{1}{l}{xh} \\
    Cherokee & chr   & Hindi & hi    & Lingala & ln    & Old French (842-ca. 1400) & fro   & Southwestern Dinka & dik   & \multicolumn{1}{l}{Yakut} & \multicolumn{1}{l}{sah} \\
    Chhattisgarhi & hne   & Hiri Motu & ho    & Lingua Franca Nova & lfn   & Old Frisian & ofs   & Spanish & es    & \multicolumn{1}{l}{Yaqui} & \multicolumn{1}{l}{yaq} \\
    Chinese & zh    & Hmong Daw & mww   & Literary Chinese & lzh   & Old Norse & non   & Standard Malay & zsm   & \multicolumn{1}{l}{Yiddish} & \multicolumn{1}{l}{yi} \\
    Choctaw & cho   & Ho    & hoc   & Lithuanian & lt    & Old Russian & orv   & Standard Moroccan Tamazight & zgh   & \multicolumn{1}{l}{Yoruba} & \multicolumn{1}{l}{yo} \\
    Church Slavic & cu    & Huastec & hus   & Liv   & liv   & Old Spanish & osp   & Sumerian & sux   & \multicolumn{1}{l}{Zarma} & \multicolumn{1}{l}{dje} \\
    Chuvash & cv    & Hungarian & hu    & Lojban & jbo   & Oriya (macrolanguage) & or    & Sundanese & su    & \multicolumn{1}{l}{Zaza} & \multicolumn{1}{l}{zza} \\
    Coptic & cop   & Hunsrik & hrx   & Lombard & lmo   & Orizaba Nahuatl & nlv   & Swabian & swg   & \multicolumn{1}{l}{Zulu} & \multicolumn{1}{l}{zu} \\
    Cornish & kw    & Hupa  & hup   & Low German & nds   & Oromo & om    & Swahili (individual language) & swh   &       &  \\
    \bottomrule
    \end{tabular}%
    }
  \caption{List of Languages. Our dataset mainly use ISO 639 series as language code. For traditional Chinese, we define ``zhtrad'' as code.}
\label{tab:lang list}%
\end{table*}%

% \section{Full Results}

% We report all the evaluation results on 86 languages, as shown in Figure~\ref{fig:detail_result_of_one2many} and Figure~\ref{fig:detail_result_of_many2one}. Lego-MT beats baselines by a large margin.

% shows the spBLEU score of different model transaltion from centric languages (en, zh, de, ar, ne, az, ceb), and  
% is the performance of translation into centric languges. On average, Lego MT outperform the competition.

% \begin{figure*}[!t]
%     \centering
%     \includegraphics[width=0.7\linewidth]{image/from_heatmap.png}
%     \caption{The One-to-Many results (spBLEU) of all models on the 86 languages of \textit{Flores-101} devtest. Each cell (row, column) represents spBLEU by translating from the row language to the column language. Lego-MT beat strong baselines by a large margin. }
%     \label{fig:detail_result_of_one2many}
% \end{figure*}

% \begin{figure*}[!t]
%     \centering
%     \includegraphics[width=0.7\linewidth]{image/to_heatmap.png}
%     \caption{The Many-to-One results (spBLEU) of all models on the 86 languages of \textit{Flores-101} devtest. Each cell (row, column) represents spBLEU by translating from the column language to the row language. Lego-MT beat strong baselines by a large margin. }
%     \label{fig:detail_result_of_many2one}
% \end{figure*}

\end{document}